\documentclass[10pt,twocolumn,letterpaper]{article}

\usepackage{cvpr}
\usepackage{times}
\usepackage{epsfig}
\usepackage{graphicx}
\usepackage{amsmath}
\usepackage{amssymb}

\usepackage[breaklinks=true,bookmarks=false]{hyperref}

\usepackage{booktabs}
\usepackage{subfigure}
\usepackage[ruled,linesnumbered]{algorithm2e}
\usepackage{algorithmic}
\usepackage{color}                      
\usepackage{multirow}

\def\mb{\mathbf}
\def\t{\intercal}
\def\op{\tt\small}

\cvprfinalcopy 

\def\cvprPaperID{****} 

\begin{document}

\title{Generating Holistic 3D Scene Abstractions for Text-based Image Retrieval}

\author{Ang Li\quad Jin Sun\quad Joe Yue-Hei Ng\quad Ruichi Yu\quad Vlad I. Morariu\quad Larry S. Davis\\
Institution for Advanced Computer Studies\\
Univesrity of Maryland, College Park, MD 20742\\
{\tt\small \{angli,jinsun,yhng,richyu,morariu,lsd\}@umiacs.umd.edu}
}

\maketitle

\begin{abstract}
Spatial relationships between objects provide important information for text-based image retrieval. As users are more likely to describe a scene from a real world perspective, using 3D spatial relationships rather than 2D relationships that assume a particular viewing direction, one of the main challenges is to infer the 3D structure that bridges images with users' text descriptions. However, direct inference of 3D structure from images requires learning from large scale annotated data. Since interactions between objects can be reduced to a limited set of atomic spatial relations in 3D, we study the possibility of inferring 3D structure from a text description rather than an image, applying physical relation models to synthesize holistic 3D abstract object layouts satisfying the spatial constraints present in a textual description. We present a generic framework for retrieving images from a textual description of a scene by matching images with these generated abstract object layouts. Images are ranked by matching object detection outputs (bounding boxes) to 2D layout candidates (also represented by bounding boxes) which are obtained by projecting the 3D scenes with sampled camera directions. We validate our approach using public indoor scene datasets and show that our method outperforms baselines built upon object occurrence histograms and learned 2D pairwise relations.
\end{abstract}

\section{Introduction}

Text-based image retrieval, dating back to the late 1970s, has evolved from a keyword-based task to a more challenging task based on natural language descriptions (e.g., sentences and paragraphs) \cite{scenegraph:cvpr2015,Kordjamshidi:2011,Rui1999}. Queries in the form of sentences rather than keywords refer to not only object categorical information but also interations, such as spatial relationships, between objects. Those relationships are usually described in the real (3D) world due to the nature of human language. Intuitively, they can be the core feature for ranking images in many application scenarios, e.g., a user searching for images that are relevant to a particular mental image of a room layout. Not surprisingly, researchers have recently increased their focus on understanding spatial relationships from text input and retrieving semantically consistent visual information \cite{scenegraph:cvpr2015,LinCVPR14,schuster-2015,zitnick-iccv13}.


Matching images with user provided spatial relations is challenging because humans naturally describe scenes in 3D while images are 2D projections of the world. Inferring 3D information from a single image is difficult. Most existing approaches learn from annotated data to map language directly to a probability distribution of pairwise relationships between object locations \cite{scenegraph:cvpr2015,LinCVPR14}. However, such a distribution is non-convex and highly non-linear in the 2D image space because the (unknown) camera view affects the bounding box configurations. Consequently, the success of 2D learning based approaches naturally depends on the size of annotated training data. Also, the learner overfits easily since annotated spatial relations have a long-tailed distribution; many valid configurations happen rarely in the real world (e.g., a desk on another desk). With pairwise relations, it is also hard to enforce the fact that all objects are viewed from the same direction in an image. This argues for a holistic model for object relationships that jointly optimizes object configurations. Motivated by this, we explore an alternative model of spatial relations that generates 3D configurations explicitly based on physics.

We explore an approach that uses physical models and complex spatial relation semantics as part of an image retrieval system  that generates 3D object layouts from text (rather than from images) and performs image retrieval by matching 2D projections of these layouts against objects detected in each database image. Our framework requires the a priori definition of a fixed set of object and spatial relation categories. Spatial relation terms are extracted from  the dependency tree of the text. Objects are modeled using cuboids and spatial relations are modeled as inequality constraints on object locations and orientations. These inequality constraints can become very complex, containing nonlinear transformations represented using first order logic. Consequently, an interval arithmetic based 3D scene solver is introduced to search for feasible 3D spatial layout solutions. Camera orientations are constrained and sampled for obtaining 2D projections of candidate scenes. Finally, images are scored and ranked by comparing object detection outputs to a sampled set of 2D reference layouts. 


Compared to 2D learning based approaches, our approach has the following advantages: (1) the mapping from language to 3D is simple since the text-based spatial constraints have a very concrete and simple meaning in 3D, simple enough to define with a few rules by hand; (2) no training data is needed to learn complex distributions over the spatial arrangement of 2D boxes given linguistic constraints (the non-linear mapping from language to 2D is handled by projective geometry) and (3) adding common sense constraints is easy when referring to physical relationships in 3D (Sec.\ \ref{sec:relationmodel}), while it is hard if these constraints are specified and learned in 2D (due to the non-linearity of projective geometry). We evaluate our approach using two public scene understanding datasets \cite{Choi:cvpr:2013,Song_2015_CVPR}. The results suggest that our approach outperforms baselines built upon object occurrence histograms and learned 2D relations.

\section{Related Work}

Text-based image retrieval has been studied for decades \cite{Rui1999}. As both computer vision and natural language processing have advanced, recent efforts have emerged that build connections between linguistic and visual information \cite{krishnavisualgenome,word2vec}. Srivastava and Salakhutdinov \cite{JMLR:v15:srivastava14b} extend Deep Boltzman Machines (DBMs) to multimodal data for learning joint representations of images and text. They apply such representations to retrieving images from text descriptions. Their model learns mappings between objects with attributes and their corresponding visual appearances; however spatial relations are not modeled.

Spatial relationships play an important role in visual understanding. Previous works make use of text-extracted spatial relations in image retrieval. Zitnick \etal \cite{zitnick-iccv13} generate and retrieve abstract cartoon images from text. Cartoon object models are pre-defined and 2D clipart images are composed according to the text. Siddiquie \etal \cite{behjat:icmr:2014} devise a multi-modal framework for retrieving images from sources including images, sketches and text by jointly considering objects, attributes and spatial relationships, and reducing all sources into 2D sketches. However, their framework handles text with only two or three objects and very limited 2D spatial relationships. Lin \etal \cite{LinCVPR14} retrieve videos from textual queries. A set of motion text is defined with visual trajectory properties and parsed into a semantic graph to to match video segments via a generalized bipartite graph matching. All these works rely on 2D spatial relations while our work is based on real world physical models of 3D scenes to retrieve semantically consistent images.

Interesting recent work on retrieving images from text is based on the scene graph representation \cite{scenegraph:cvpr2015,schuster-2015}. A scene graph is a graph-based representation which encodes objects, attributes and object relations. In  Johnson \etal \cite{scenegraph:cvpr2015}, text input is converted to a scene graph by a human and a CRF model is used to match scene graphs to images by encoding global spatial relations of objects rather than only pairwise relations. Their approach requires learning spatial relations from annotated image data. Our work differs in that we take a generative perspective and inject physical relation models and human knowledge into the retrieval system without the requirement of large-scale data annotation.

Many existing works utilize 3D geometry in vision tasks such as object recognition \cite{Hoiem2008}, image matching \cite{li14geolocation}, object detection \cite{xiang-iccv13, Zia2015}, \etc. However, to the best of our knowledge, the use of 3D geometry in relating images with language has not been exploited. While inferring the 3D structure from a single image is challenging and complicated in vision \cite{Choi:cvpr:2013,eigen2015,book2011Hoiem,pero-cvpr12,del-pero-2013}, the problem of
rendering scenes from text is of interest in the graphics community.
The wordseye system \cite{wordseye} renders scenes from text with given 3D object models. Chang \etal \cite{chang-emnlp14} generates 3D scenes from text by incorporating the spatial knowledge learned from data. In addition, some recent works cast computer vision as inverse graphics and try to incorporate computer graphics elements into visual understanding systems \cite{kulkarni2014deep,DBLP:journals/corr/KulkarniMKT14,galileo}. Our work also involves scene generation. However, our purpose is to retrieve similar images based on bounding boxes , which can be efficiently computed using off-the-shelf software during a database indexing step,  so real object models are not required, although better scene generation could potentially improve image retrieval accuracy. 


\section{Preliminary: Interval Analysis}
Our approach involves finding feasible solutions to a mathematical program where the variables are object coordinates and orientations, and the constraints are inequalities translated from user descriptions. Since small placement pertubations usually do not affect the fullfilment of constraints, feasible variables can naturally be represented by a set of intervals (any value within the interval is feasible).

\begin{figure*}[!t]
\centering
\includegraphics[width=\textwidth]{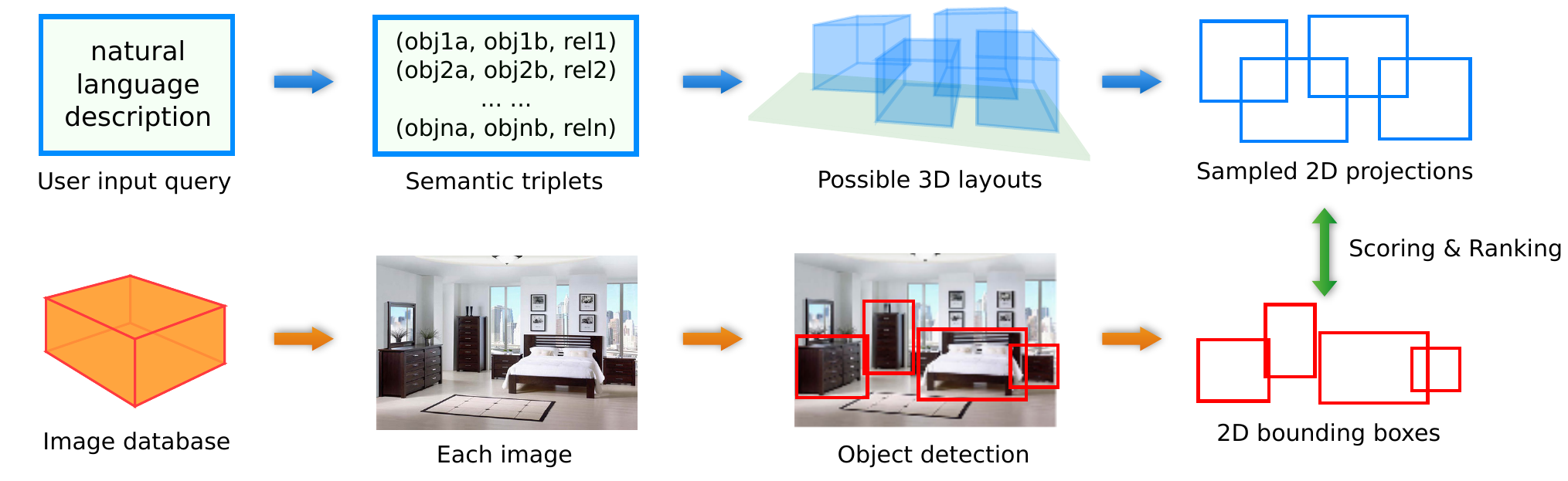}
\caption{Framework overview: a textual description of the visual scene is parsed into semantic triplets which are used for solving feasible 3D layouts and their 2D projections as \emph{reference configurations}. An object detector runs over each database image and generates a 2D bounding box layout, to be matched to reference configurations. All database images are ranked according to their configuration scores.}\label{overview}
\end{figure*}

Interval analysis represents each variable by its feasible interval, e.g., $[l, u]$ (with lower bound $l$ and upper bound $u$) and the goal is to find the bound for each dimension that satisfies all constraints \cite{Snyder:1992:IAC}. When an interval does not satisfy all the constraints, it is split into smaller intervals and evaluated recursively. Arithmetic operators are defined in terms of intervals, e.g.,
\vspace{-0.5\topsep}
  \begin{itemize}
  \itemsep -0.3em
  \item \textit{addition}: $[l_1, u_1]+[l_2,u_2]=[l_1+l_2,u_1+u_2]$;
  \item \textit{subtraction}: $[l_1, u_1]-[l_2,u_2]=[l_1-u_2,u_1-l_2]$;
  \item \textit{comparison}: 
  $[l_1, u_1] < [l_2, u_2]$ equals $[0, 0]$ if $u_2\le l_1$ (definitely false); equals $[1, 1]$ if $u_1<l_2$ (definitely true); equals $[0, 1]$ otherwise (maybe true).
  \end{itemize}\vskip -0.5\topsep
  The fulfillment of a constraint can be represented by any of the   three logical intervals, \ie, $[0, 0], [1, 1], [0, 1]$.

\section{Our Approach}
The proposed framework, as illustrated in Fig. \ref{overview}, consists of several modules.
First, the input text is parsed into a set of semantic triplets of object names and their spatial relationships.
Second, the semantic triplets are used to solve possible 3D layouts of objects along with sampled camera locations and orientations.
The 2D projections of the 3D scenes are used for generating 2D bounding boxes of objects, which we call {\it reference configurations}.
Finally, the reference configurations are matched to the detected bounding boxes in each database image to score and rank according to their configuration similarity.

\subsection{Text Parsing}

The text parsing module translates text into a set of semantic triplets which encode the information about two object instances and their spatial interactions. How to robustly extract relations from text is still an open research problem in natural language processing \cite{Kordjamshidi:2011}, which is beyond the scope of this paper. For our application, a simple rule-based pattern matching works sufficiently well, requiring a pre-defined dictionary of object and spatial relation categories. 
A text example and its parsing output is shown in Table \ref{tab:textparsing}.

The input text is processed by the Stanford CoreNLP library \cite{manning-EtAl:2014:P14-5} with part-of-speech tagging and dependency tree. We implement a rule-based approach to extract spatial relations  (such as \textit{on}, \textit{under}, \textit{in front of}, \textit{behind}, \textit{above}, etc.) from the dependency tree and compose its corresponding semantic triplet representation {\it (target object, reference object, relation)}. The co-reference module in the CoreNLP library is used to aggregate multiple noun occurrences that correspond to the same object instance. Each object reference is represented by its category name and a unique ID within the category, e.g. {\op sofa-0} and {\op dining-table-2}. 

Natural objects are usually composed of multiple sub-objects and there are often cases when a sub-object is referenced instead of the whole object. A bed, for instance, has its head and rear. And a chair has its back and seat. We take sub-objects into consideration and represent any sub-object reference by its object category name,  unique in-category ID  and sub-object name, e.g. ``the rear of the bed'' is represented as {\op bed-0:rear} if the ID is 0. 

Besides object categories and spatial relationships, we also consider the count of each object, e.g. three chairs, two monitors, etc. The parser maintains a list of object ID and their counts. If the count of {\op chair-0} is 3, then the parser will expand {\op chair-0} to a set of three instances {\op\{chair-0-0,chair-0-1,chair-0-2\}} in the outputs. 

\begin{table}[!tb]
\centering
\small
\setlength\tabcolsep{2.5pt}
\begin{tabular}{cl}
\toprule
\bf \#&\bf Sentence $\rightarrow$ (object-1, object-2, relation)\\ \midrule
1&A picture is above a bed.\\
&\op (picture-0, bed-0, above) \\ 
2&A night stand is on the right side of the head of the bed.\\
&\op (night-stand-0, bed-0:head, right) \\ 
3&A lamp is on the night stand. \\
&\op (lamp-0, night-stand-0, on) \\ 
4&Another picture is above the lamp.\\
&\op (picture-1, lamp-0, above)\\ 
5&A dresser is on the left side of the head of the bed.\\
&\op (dresser-0, bed-0:head, left)\\\bottomrule
\end{tabular}
\vskip 3pt
\caption{Semantic triplet parsing from an example query}
\label{tab:textparsing}
\end{table}

  \subsection{3D Abstract Scene Generation}
  The 3D abstract scene generation module is the central component in our image retrieval framework; it takes as input semantic triplets and generates a set of sampled possible 3D object layouts. We describe below the three core components of the scene generator: the cuboid based object model, the spatial relation model and the 3D scene solver. 
  
  \subsubsection{Cuboid based object model} 
The basic \textit{cuboid representation} of an object is $\mb C=(l_x, l_y, l_z, z_s)$ where $(l_x, l_y, l_z)$ is the size of the cuboid that bounds the object in $x,y,z$ directions respectively and $z_s$ is the $z$-coordinate of the supporting surface of the object. We mostly use regular sizes but also set different sizes for objects with attributes such as {\op long-desk}, {\op triple-sofa}, etc. The supporting surface is usually the top face of the object cuboid, but it can sometimes be located elsewhere with respect to the cuboid, e.g., for a chair it is in the middle of the cuboid. Spatial relations such as {\op on} and {\op under} are modeled with respect to the surface of the object. Most of the objects can be modeled using this cuboid representation such as {\op garbage-bin}, {\op picture}, {\op night-stand}, etc.
  
\begin{figure}[!t]
    \centering
    \subfigure[]{
  \includegraphics[height=1.7cm]{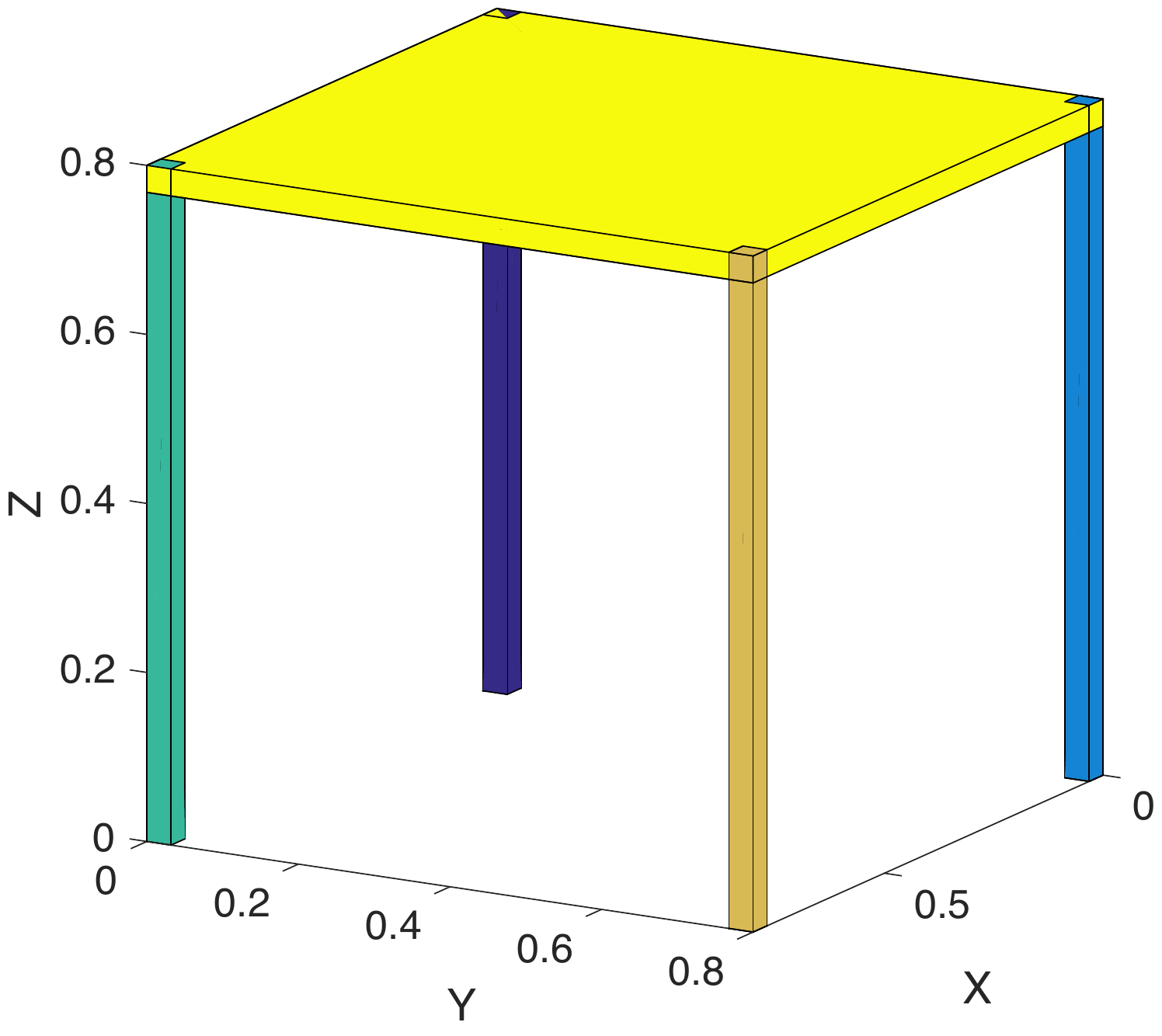}
  }\quad\subfigure[]{
  \includegraphics[height=1.9cm]{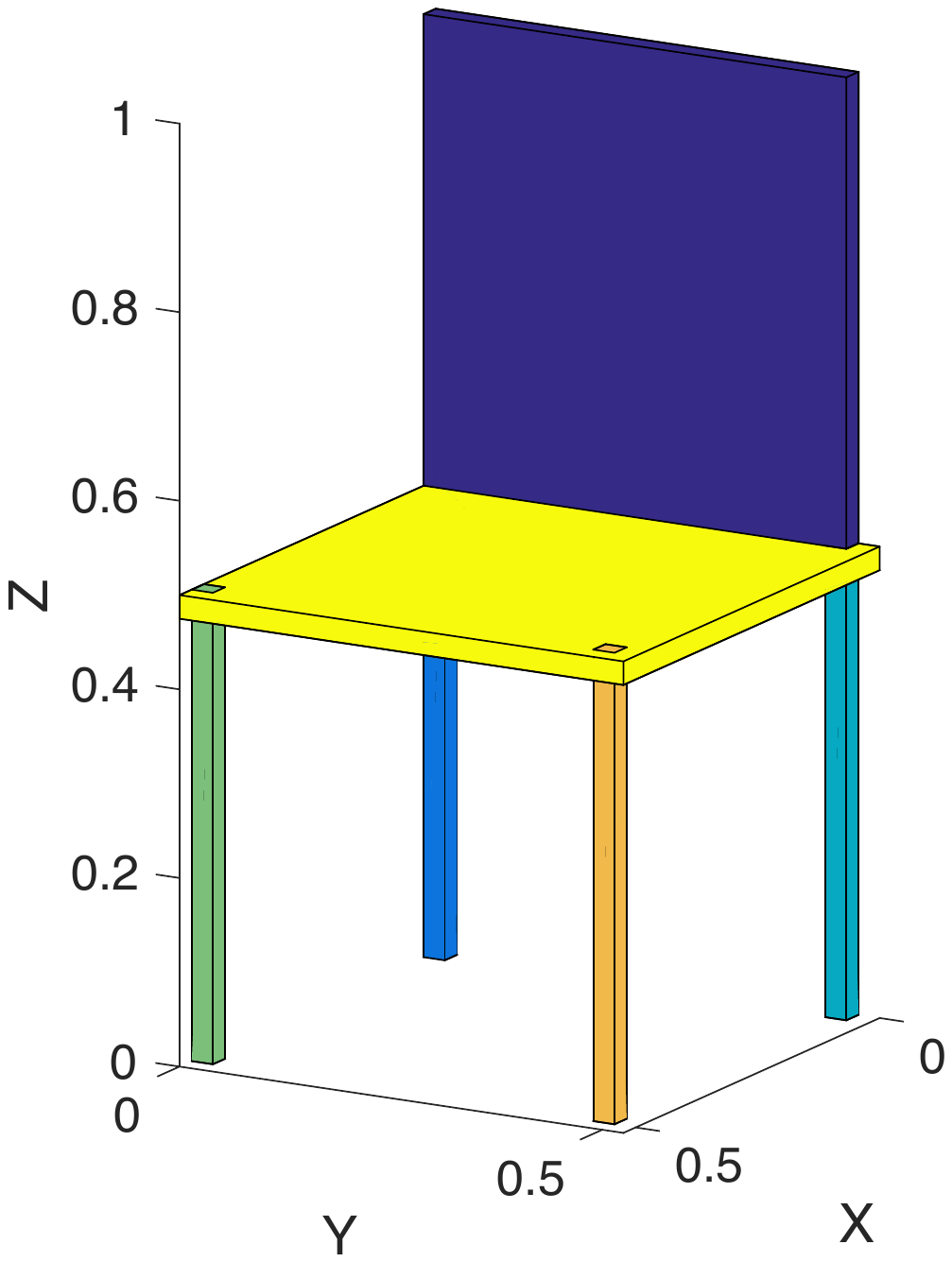}
  }\quad\subfigure[]{
  \includegraphics[height=1.9cm]{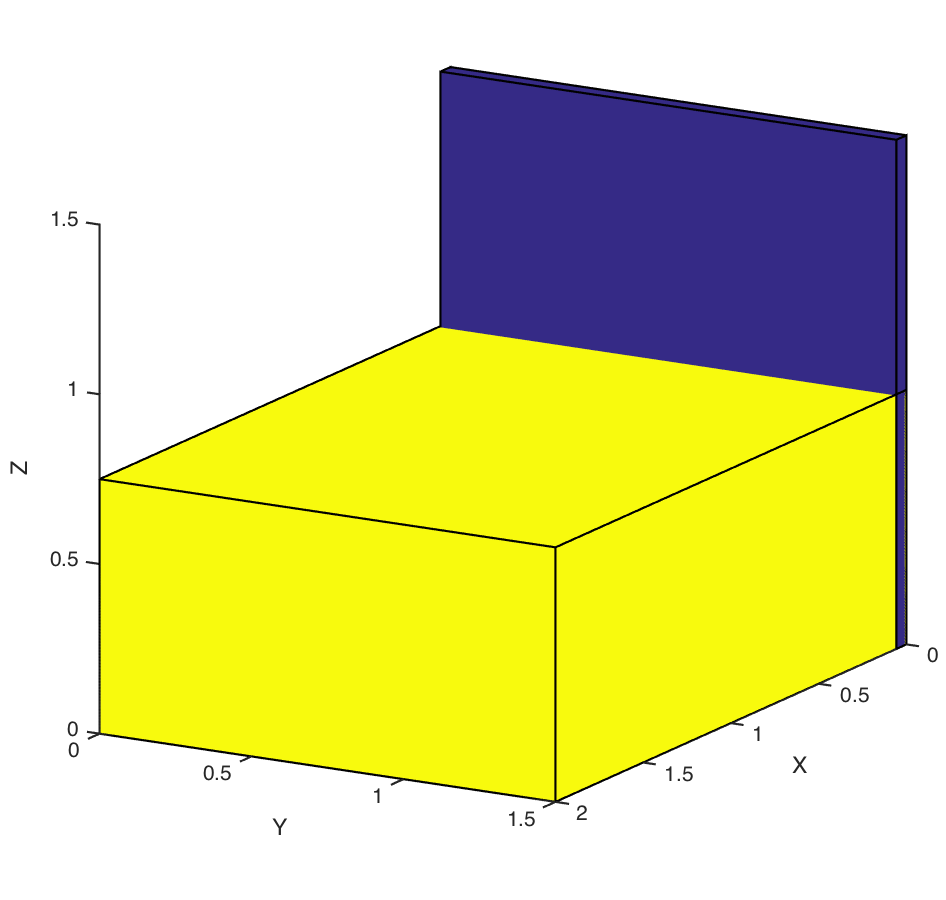}
  }\quad\subfigure[]{
  \includegraphics[height=1.3cm]{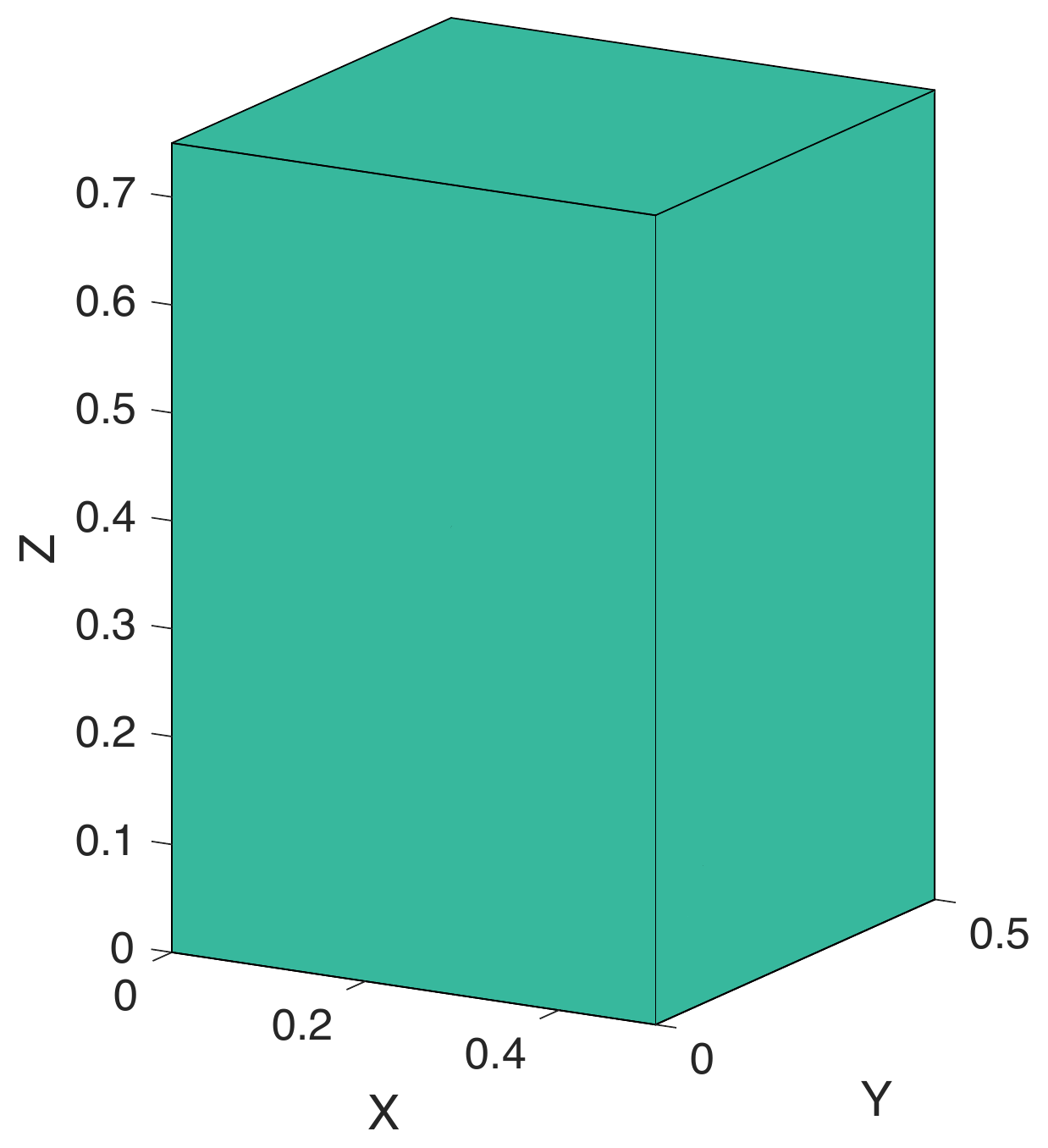}
  }
  \caption{Sample cuboid based object representations: (a) table (b) chair (c) bed (d) night-stand. Different colors represent different sub-objects. The night stand (d) is represented by a single cuboid.}
  \label{fig:objmodels}
\end{figure}

However, the single cuboid representation is not sufficient for some object categories such as {\op chair} and {\op desk} since the under-surface area is empty. Considering the fact that most objects can be easily decomposed into smaller sub-objects, we represent these object categories as the union of a set of cuboids, which we call a \textit{cuboid set representation}. Each sub-cuboid corresponds to a sub-object and is considered a simple object, whose top face is the supporting surface. The $k$-th sub-cuboid is represented by $\mb S^k=(d_x^k, d_y^k, d_z^k, l_x^k, l_y^k, l_z^k)$ where $(d_x^k, d_y^k, d_z^k)$ is the offset from the lowest point of the sub-cuboid to the lowest point of the original object, and $(l_x^k, l_y^k, l_z^k)$ is the size of the sub-cuboid. The sub-cuboid parameters $\mb S^k$ are computed as functions of the original object parameters $\mb C$. Four sampled cuboid based object models are visualized in Fig. \ref{fig:objmodels}.

  \subsubsection{Spatial relation model}\label{sec:relationmodel}
  The spatial location and orientation of each object is represented as $\mb X=(x, y, z, d)$ where $(x, y, z)$ is the lowest point of the object cuboid and $d$ is its orientation. The object rotation is  around the $z$-axis.
  
  \textbf{Atomic relations.}
  We model 8 basic spatial relations using the following mathematical expressions. Given the object pose and its size, the lowest point $\mb p=(x_p, y_p, z_p)^\t$ and highest point $\mb q=(x_q, y_q, z_q)^\t$ of the object cuboid can be computed by rotating the object models w.r.t. the object orientation such that
  \begin{align}
  \mb p&=\mb R_d\left[-\frac{l_x}{2}, -\frac{l_y}{2}, -\frac{l_z}{2}\right]^\t+\left[x+\frac{l_x}{2},y+\frac{l_y}{2},z+\frac{l_z}{2}\right]^\t,\nonumber\\
  \mb q&= \mb R_d\left[\frac{l_x}{2}, \frac{l_y}{2}, \frac{l_z}{2}\right]^\t+\left[x+\frac{l_x}{2},y+\frac{l_y}{2},z+\frac{l_z}{2}\right]^\t
  \end{align}
   where $\mb R_d$ is the $z$-axis rotation matrix w.r.t. to orientation $d$. So an object can be represented using tuple $(\mb p, \mb q, d)$.
Letting the cuboid of object-1 be $\mb O_1(\mb p_1, \mb q_1, d_1)$ with support surface $z_{s1}$ and the cuboid of object-2 be $\mb O_2(\mb p_2, \mb q_2, d_2)$ with support surface $z_{s2}$, we define 8 atomic relations as\vspace{-0.5\topsep}
  \begin{itemize}
  \itemsep 0em
  \item {\op near}: $\mb O_1 \cap (\mb p_2 - d_\text{near}\mb e_{d_2}, \mb q_2 + d_\text{near}\mb e_{d_2}, d_2)\not=\emptyset$;
      \item {\op on}: $z_{p1}=z_{s2}\wedge\frac{\mb p_1+\mb q_1}{2}\in_{xy}\mb O_2$;
      \item {\op above}: $z_{q2}+d_\text{min-above}\le z_{p1}\le z_{q2}+d_\text{max-above}\wedge\frac{\mb p_1+\mb q_1}{2}\in_{xy}\mb O_2$;
      \item {\op under}: $z_{s1}<z_{s2}\wedge \mb O_1\cap_{xy} \mb O_2\not=\emptyset$;
            \item {\op behind}: $\max(\mb u_{d_2}^\t\mb p_1,\mb u_{d_2}^\t\mb q_1)\le \min(\mb u_{d_2}^\t\mb p_2, \mb u_{d_2}^\t\mb q_2)$;
    \item {\op front}: $\min(\mb u_{d_2}^\t\mb p_1,\mb u_{d_2}^\t\mb q_1)\ge \max(\mb u_{d_2}^\t\mb p_2, \mb u_{d_2}^\t\mb q_2)$;
  \item {\op on-left}: $\min(\mb u_{d_2-\pi/2}^\t\mb p_1,\mb u_{d_2-\pi/2}^\t\mb q_1)\\~~~~~~~~~~~~~~~~~~~~\ge \max(\mb u_{d_2-\pi/2}^\t\mb p_2, \mb u_{d_2-\pi/2}^\t\mb q_2)$;
    \item {\op on-right}: $\max(\mb u_{d_2-\pi/2}^\t\mb p_1,\mb u_{d_2-\pi/2}^\t\mb q_1)\\~~~~~~~~~~~~~~~~~~~~\le \min(\mb u_{d_2-\pi/2}^\t\mb p_2, \mb u_{d_2-\pi/2}^\t\mb q_2)$;
      \end{itemize}\vspace{-0.5\topsep}
      where $d_\text{near}, d_\text{min-above}, d_\text{max-above}$ are distance thresholds, $\mb p \in_{xy} \mb C$ means point $\mb p$ is inside the cuboid $\mb C$ on the $x$-$y$ plane, $\cap$ represents the intersection of two cuboids and $\cap_{xy}$ the intersection of two cuboids on the $x$-$y$ plane, and $\mb u_\theta=(\cos\theta, \sin\theta, 0)^\t$ is a unit direction vector and $\mb e_\theta=(\cos\theta-\sin\theta, \sin\theta+\cos\theta, 1)^\t$ is a vector that enlarges the effective object cuboid.

  \textbf{Composite relations.} In natural language, there are far more spatial relation descriptions than the above mentioned 8 relations. However, most of the spatial relations can be defined based on the 8 atomic relations. Two examples are\vspace{-0.5\topsep}
  \begin{itemize}
  \itemsep 0em
    \item {\op next-to}: $\text{\op on-left}(\mb O_1, \mb O_2)\vee\texttt{\op on-right}(\mb O_1,\mb O_2)$;
      \item \texttt{\op side-by-side}: $d_1=d_2\wedge\texttt{\op near}(\mb O_1, \mb O_2)$;
  \end{itemize}\vspace{-0.5\topsep}
  In addition, another relation is modeled which is usually used for a set of multiple instances $\{\mb O_1, \mb O_2, \ldots, \mb O_k\}$ of the same object category, i.e.,\vspace{-0.5\topsep}
  \begin{itemize}
  \item {\op in-a-row}: $d_i=d_{i+1}\wedge\texttt{\op on-right}(\mb O_i, \mb O_{i+1}),\ \forall i$;
  \end{itemize}\vspace{-0.4\topsep}
  
    \textbf{Group relations.} If an object reference has a count more than 1, then all of its instances form a group, which often interacts with other objects as an entirety. If a group of $k$ instances occurs in the triplet as the target, we create $k$ new triplets with the same reference and relation. If the group occurs as the reference, then we create a new virtual object whose cuboid is bounded by all of its instances.
    
  
  \textbf{Prior constraints.} An effective way to reduce the search space is to incorporate common sense and reasonable assumptions into the constraints. First, we make the following assumptions: (a) the room has two walls ($x=0$ and $y=0$); (b) the text description is coherent, i.e., the objects in each semantic triplet are close to each other; (c) objects are usually oriented along $x$-axis or $y$-axis directions. Second, no pair of objects overlap with each other, i.e.,\vspace{-0.5\topsep}
  \begin{itemize}
  \item \texttt{\op exclusive}: $\mb S_i^v\cap\mb S_j^w=\emptyset\forall i, j, v, w $
  \end{itemize}\vspace{-0.5\topsep}
  where $\mb S_i^v$ is the $v$-th component (sub-cuboid) of the $i$-th object.  Many other constraints are related with object properties: (a) picture, door, mirror are on the wall, i.e. $x=0\vee y=0$;
(b) for relation next-to, in-a-row, side-by-side, if either reference or target is against the wall, the other ones are also against the wall and they should also have the same orientation; (c) bed, night-stand, sink are against the wall; (d) bed, night-stand, sofa are on the ground.
 
  \subsubsection{3D scene solver}
  Let $\mb X=\{x_1,y_1,z_1,d_1,\ldots,x_n,y_n,z_n,d_n\}\in \mathbb{R}^{4n}$ be a \textit{layout state} representing the locations and orientations of all objects. We construct \textit{constraint function} $F: \mathbb{R}^{4n}\rightarrow \{0, 1\}$ which evaluates all prior constraints and relational constraints. The goal is to find the feasible solution set $\mb S$ such that $F(\mb X)=1$ for all $\mb X\in\mb S$. 

   Our solver is based on interval analysis \cite{Snyder:1992:IAC} where any variable is represented by an interval (an uncertain value) instead of a certain value. We use a vector of size 2 to represent an interval, i.e., a lower bound and an upper bound. Under interval analysis, the domain of layout states becomes $\mathbb{R}^{4n\times 2}$ and the constraint function becomes $F: \mathbb{R}^{4n\times2}\rightarrow \{[0,0],[0,1],[1,1]\}$. Starting with a candidate queue containing an initial interval layout state $\{\mb X_0\}$, our solver examines the candidate states one at a time. For each state $\mb X_i\in\mathbb{R}^{4n\times 2}$, if $F(\mb X_i)=[1, 1]$, then $\mb X_i$ is feasible and appended to the solution set. If the constraint fullfillment is undecidable, i.e., $F(\mb X_i)=[0,1]$, then $\mb X_i$ is divided into two equally sized intervals by splitting the variable with the largest uncertainty. The two new states are appended to the candidate queue. Otherwise, $F(\mb X_i)=[0, 0]$ and no feasible solution is within the space bounded by $\mb X_i$. In the end, any layout in the solution set is guaranteed to meet all constraints. An advantage of the method is that it does not require computing the gradient of constraint $F$.  The pseudo-code  is shown in Algorithm \ref{algo:ia}.
   
  \textbf{Interval shrinkage.} The original interval analysis does not make full use of equality constraints, e.g., when a variable is constrained to equal another variable, it becomes redundant to divide both of their intervals since one can be directly computed based on the other. In addition, many spatial relations are transitive, e.g., if object A is in front of object B and B is in front of C, then A is likely to be in front of C but with a larger distance. Such inferred constraint can benefit the solver with a better pruning power. Based on these observations, we develop the interval shrinkage operation which pre-computes lower bound matrices $\mb L^x,\mb L^y,\mb L^z\in\mathbb R^{n\times n}$ and upper bound matrices $\mb U^x, \mb U^y, \mb U^z\in\mathbb R^{n\times n}$ for pairwise coordinate differences, i.e., $L^x_{i,j}\le x_i-x_j\le U^x_{i,j}\wedge L^y_{i,j}\le y_i-y_j\le U^y_{i,j}\wedge L^z_{i,j}\le z_i-z_j\le U^z_{i,j}$. The bound matrices are initialized using the original constraints and updated once we find $L^*_{i,j}< L^*_{i,k}+L^*_{k,j}$ or $U^*_{i,j}< U^*_{i,k}+U^*_{k,j}$ ($*\in \{x, y, z\}$). Before evaluating each candidate interval layout state, we shrink its variables according to the bound matrices, e.g., $\mb x_i^\text{shrink}=\cap_j[x_j+L^x_{i,j}, x_j+U^x_{i,j}]\cap \mb x_i $ where $\mb x_i$ is the interval of variable $x_i$ and $\mb x_i^\text{shrink}$ is the interval after shrinkage.
  
 \begin{algorithm}[!t] \small
 \KwData{Initial bounds $\mb X_0=[\mb x_1, \mb y_1, \mb z_1, \mb d_1, \ldots, \mb x_n, \mb y_n, \mb z_n, \mb d_n]\in\mathbb R^{4n\times 2}$}
 \KwData{Constraint $F:\mathbb R^{4n\times 2}\rightarrow \{[0,0], [0,1], [1,1]\}$}
 \KwResult{Feasible regions (or solution set) $\mb S$}
 initialization: $\mb S=\emptyset, \mb Q=\{\mb X_0\}$\;
 \While{$\mb Q\not=\emptyset$}{
  read the first interval: $\mb X_i$ = $\mb Q$.{front}()\;
  remove the first interval: $\mb Q$.{pop}()\;
  interval shrinkage: $\mb X_i$ = {shrinkage}($\mb X_i$)\;
  
   \uIf{F($\mb X_i$)=$[0, 0]$}{
     $\mb X_i$ is not feasible\;
  }
  \uElseIf{F($\mb X_i$)=$[1, 1]$}{
  	$\mb X_i$ is feasible: $\mb S$.\text{append}($\mb X_i$)\;
  }
  \uElseIf{$\max_k|X_{ik}.{\max}-X_{ik}.{\min}|> tol$}{
  	$k=\arg\max_k|X_{ik}.{\max}-X_{ik}.{\min}|$\;
  	half split $k$-th dimension of $\mb X_i$ into $\mb X_i^{(1)}$ and $\mb X_i^{(2)}$\;
  	$\mb Q$.append($\mb X_i^{(1)}$)\;
  	$\mb Q$.append($\mb X_i^{(2)}$)\;
  }
 }
 \Return $\mb S$\;
 \caption{3D scene solver}\label{algo:ia}
\end{algorithm}

  \begin{figure*}[!t]
        \centering
  \subfigure[]{
  \includegraphics[width=0.3\textwidth]{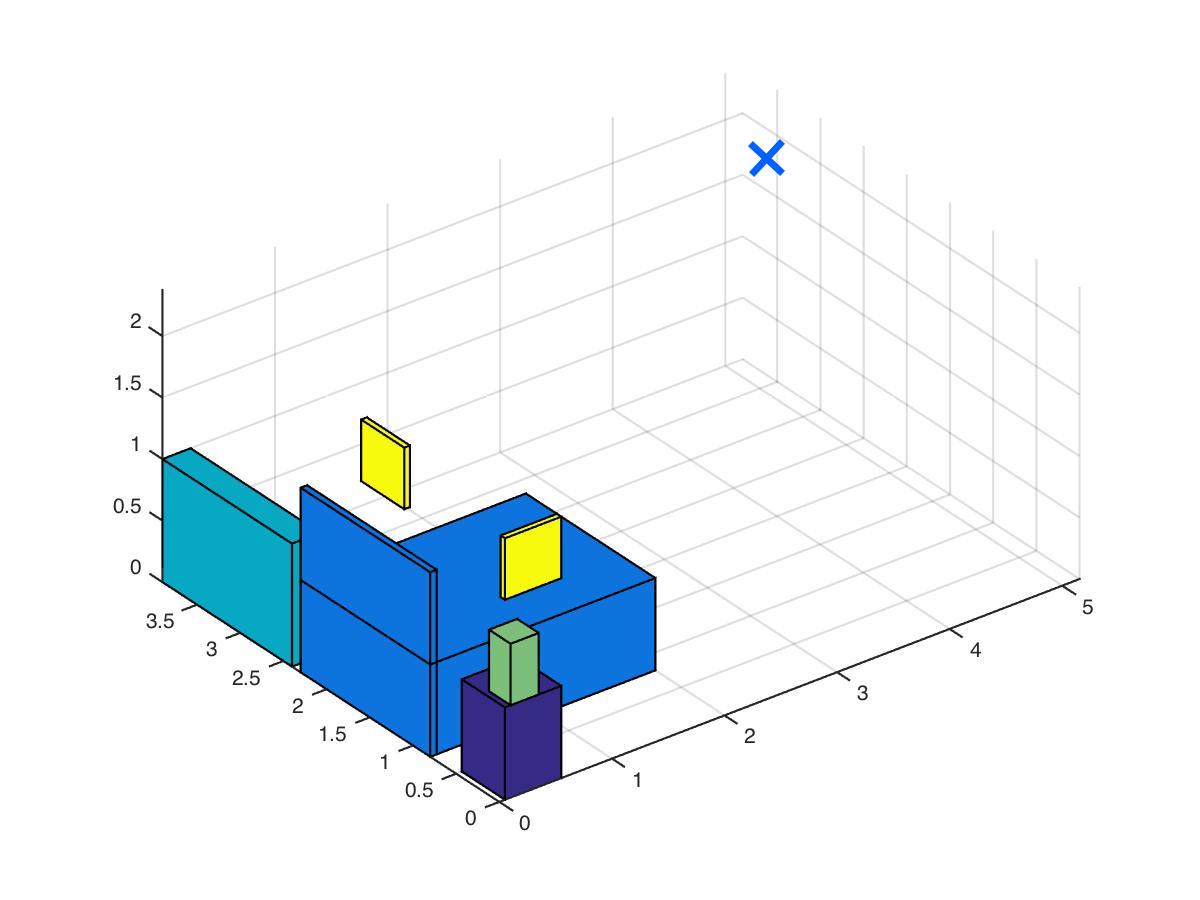}
  }\subfigure[]{
  \includegraphics[width=0.3\textwidth]{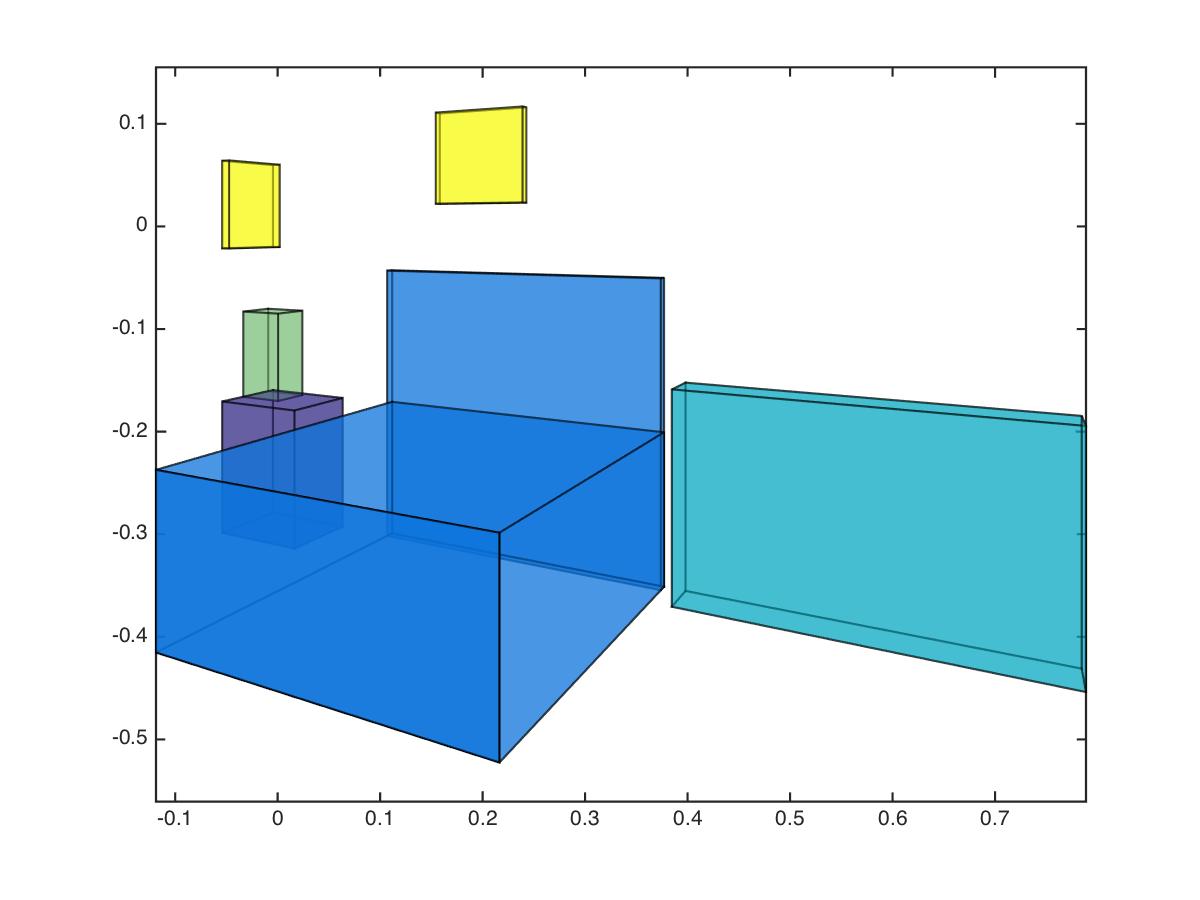}
  }\subfigure[]{
  \includegraphics[width=0.3\textwidth]{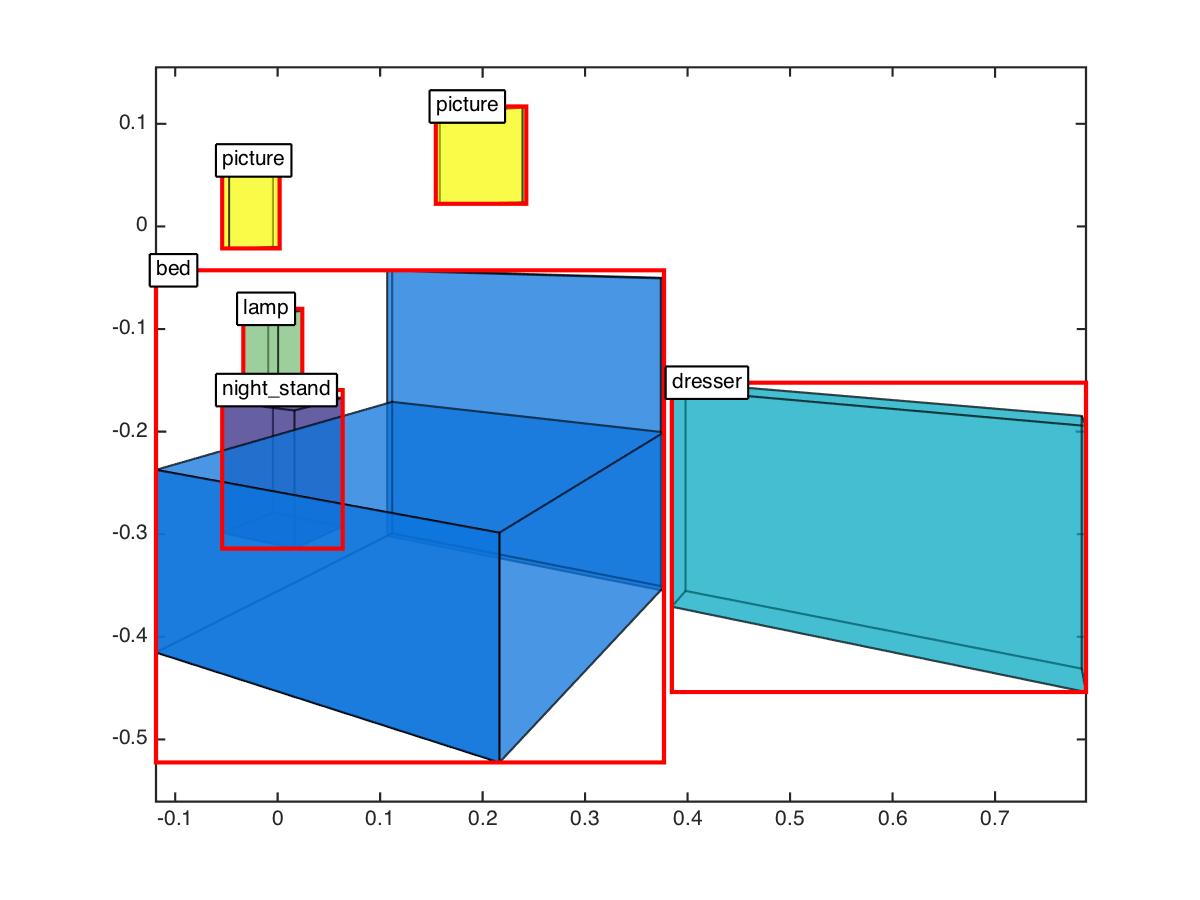}
  }
  \caption{The generated scene geometry for the query in Table \ref{tab:textparsing}: (a) a sampled 3D layout with the sampled camera location (a blue cross in the figure), (b) 2D projections of the object cuboids and (c) 2D bounding boxes of the objects.}\label{fig:solver}
\end{figure*}

\textbf{Early stopping.} The feasible solution space can be large if the input constraints are weak. Since we sample $K$ layouts in our framework for subsequent image matching, the 3D scene solver stops when at least $K$ layouts are found. The sampling behaviour is achieved by implementing the candidate queue with Knuth shuffling, i.e., each time after appending a new element, the queue randomly pick an element and swaps it with the new element.

The problem is a combinatorial optimization which is NP-hard and interval analysis is essentially a breadth first search with pruning. As a result, the algorithm has no time limit guarantee. However, with interval shrinkage and early stopping, our algorithm is able to solve most queries in a reasonable amount of time. Without interval shrinkage, our \textsc{matlab} implementation can not find a solution for the query in Table \ref{tab:textparsing} within 10 minutes, while it returns 5 solutions with only 6 seconds using the shrinkage operation.

  \subsection{Image Retrieval}
To compare a query with image bounding boxes, we first sample feasible 3D layouts and potential camera locations and orientations to produce reasonable 2D projections of objects and then compute their bounding boxes. The whole image database is scored and ranked according to the similarity between bounding boxes detected by object detectors and those from sampled 2D layouts.


\textbf{3D layout sampling.}  The 3D solver finds (continuous) interval solutions for 3D object coordinates; any solution within such intervals is feasible. However, the solutions within an interval are redundant; those object locations shift in tiny distances. So we sample only one layout within each interval, which results in a set of representative feasible 3D layouts. We further sample a few 3D layouts from this feasible set in order to generate their 2D projections.

\textbf{2D layout projections.}  For each layout, we sample camera locations and orientations to obtain 2D projections which allows matching images under multiple views. Object bounding boxes are computed according to the 2D projections. Since we solve for scale and translation for each image individually during matching, in this step we only consider a canonical camera. Some heuristics are used for sampling camera locations and orientations. First, the camera always faces the objects and should be neither too close nor too far, so we sample its location from 5-10 meters from the origin. Second, the camera should not be located behind the wall, so the coordinates are positive. Third, when an object is on the wall, the camera direction should be within 60 degree offset from the object orientation. We assume the camera is 1.7 meters above the ground and situated horizontally. Fig. \ref{fig:solver} shows an example of 3D layout, 2D projections and 2D bounding boxes for the query in Table \ref{tab:textparsing}.


  \textbf{2D layout similarity.} Both detection outputs and 2D reference layouts can be represented by $\{\mb b_i, c_i\}$ where $\mb b_i$ is the 2D box of the $i$-th object and $c_i$ is its category. Let $\{\mb b_i, c_i\}$ be a 2D reference layout and $\{\mb b'_i, c'_i\}$ be the detected  boxes. Since scaling and translation are left as free variables, the bounding box matching involves optimizing\vskip -1.5\topsep
  \begin{align}
  \max_{s,t, \mb a} \sum_ip(\mb b'_{a_i})\cdot \textsc{iou}(s\mb b_i+t, \mb b'_{a_i}),\quad s.t.\  c_i=c'_{a_i},\label{layout-match}
  \end{align}\vskip -\topsep
  \noindent where $p(\mb b'_k)$ is the detection confidence, \textsc{iou} is intersection-over-union and assignment vector $\mb a$ indicates the correspondence between two sets of bounding boxes. In our experiment, we evaluate two versions: (a) the \textit{hard} version uses a threshold on detection outputs and uniform $p(\mb b'_k)$ and (b) the \textit{soft} version makes $p(\mb b'_k)$ equal to the detection score.
  We use a sliding window to find the best matched transformation and assignment. Specifically, we uniformly sample $5$ scale factors from $0.5$ to $1$ \wrt the image space and search with a $10$-pixel stride. We use a greedy strategy to compute assignments and scores (Eq. \ref{layout-match}). The score for a query is computed as the highest score among the scores of all its sampled 2D layouts.

\section{Experiments}
We validate our approach using two indoor scene datasets (SUN RGB-D \cite{Song_2015_CVPR} and 3DGP \cite{Choi:cvpr:2013}). Although the original goal of the two datasets is not text-based image retrieval, both contain groundtruth object bounding boxes which enables evaluation in our image retrieval setting. We compare 3 baselines built upon object occurrence histogram and 2D spatial relation based scene graph matching. 

\subsection{Setup}

\textbf{Baseline (H).} The first baseline is based on the histogram of object occurrences.  Specifically, both the image and text are converted to a histogram representation, i.e., a vector $\mb x=\{x_1, x_2, \ldots, x_N\}$, where $x_i$ is the number of occurrences of the $i$-th object category. The similarity between occurrence histograms is measured by $\ell^1$ distance.


\textbf{Baseline (2D).} The second baseline is based on learned object relations in 2D image space. Specifically, the baseline learns a bounding box distribution of the first object \wrt the second object box (normalized in both $x$ and $y$ coordinates). We have all eight atomic relations annotated in 1,000 images in the training set of SUN RGB-D dataset and use IOU-based nearest neighbor (IOU-NN) classifier to score for each test image the spatial relationships between object pairs. Following \cite{spice2016}, we convert the text to a simplified scene graph that maps all instances of an object category into a single node, and assign the count of each relation as an attribute of the corresponding edge. An image scene graph with relation probabilities on edges can be constructed for each test image by using the IOU-NN relation classifier upon each pair of detected object instances. To measure the similarity between text scene graph and image scene graph, we sum for each edge $(u,v,r)$ in the text scene graph the top $k_{u,v,r}$ corresponding relation scores in the image scene graph, where $k_{u,v,r}$ is the count of the relation $r$ between object categories $u$ and $v$ in text scene graph.

\textbf{Baseline (CNN).} The third baseline replaces the IOU-NN relation classifier in Baseline 2D with a Convolutional Neural Network (CNN). Following \cite{lu2016visual}, we finetune the pretrained VGG-19 \cite{vggnet} to predict predicates from cropped union image regions of the two objects. The word2vec vectors of the two objects are concatenated with the response of layer \textit{fc7}. We backpropagate through the whole network with initial learning rate $0.001$ for $90$ epochs.

\textbf{Evaluation metric.} We evaluate different approaches to retrieving indoor images from text descriptions by measuring the percentage of queries (recall) at least one of whose ground truth images are retrieved within top $k$ ranked images (\textit{R@$k$}). The median rank (median of the ranks of all ground truths) is used as a global measurement.

 \textbf{Parameter selection.} We set the room size to be $5m\times5m\times5m$. $d_\text{near}=0.5m,d_\text{min-above}=0.25m,d_\text{max-above}=0.5m$. The tolerance in 3D scene solver is $0.2m$ because $20cm$ replacement of objects is unlikely to change the constraint fulfillment. We sample 5 reference layouts per query and 1 camera view per layout unless otherwise specified.

\subsection{SUN RGB-D dataset with R-CNN detectors}
SUN RGB-D Dataset \cite{Song_2015_CVPR} is a recent dataset for scene understanding which contains $10,335$ RGBD images. We use only the RGB images without depth information. We follow the same protocol as \cite{Song_2015_CVPR} by using $5,285$ images for training the detectors and the remaining $5,050$ images as the evaluation set. We annotated text queries for $150$ sampled test images. SUN RGB-D contains various objects and complex spatial relations. We choose $19$ object categories in our evaluation:  \{\textit{bed, chair, cabinet, sofa, table, door, picture, desk, dresser, pillow, mirror, tv, box, whiteboard,
night\_stand, sink, lamp, garbage\_bin, monitor}\}, which contains not only objects on the floor but also those off the ground or on the wall such as \textit{picture} and \textit{mirror}. 


We use the $5,285$ training images and their ground truth object bounding boxes to train Fast R-CNN \cite{girshick15fastrcnn} detectors for the 19 object categories. The R-CNN approach is built upon object proposals; non-maximum suppression is not used in postprocessing. For each test image, R-CNN detectors generate probability-like scores for all object categories on each object proposal bounding box. The category with the highest score is chosen as the bounding box category and its score is used as the bounding box confidence. 

\begin{table}[!tb]
\centering
\resizebox{\linewidth}{!}{
\begin{tabular}{lccccc}
\toprule
&\bf R@1&\bf R@10&\bf R@50&\bf R@100&\bf R@500\\
\midrule
Baseline H &1.3	&4.0	&14.0	&20.0	&43.3\\
Baseline 2D	&2.7	&15.3	&35.3	&44.0	&64.0\\
Baseline CNN & 	2.7	&	16.7 & 30.7 & 36.0 & 63.3\\
Ours hard[5,1]	&3.9	&16.4	&31.7	&42.3	&71.7\\
Ours soft[5,1]	&4.5	&16.7	&34.0	&46.4	&76.0\\
Ours soft[5,5]	& \underline{4.9}	& \underline{18.7}	&\underline{37.9}	&\underline{48.1}	&\underline{76.9}\\
Ours soft[5,5] + 2D 		&\bf 8.7	&\bf 21.6	&\bf40.5	&\bf50.7	&\bf77.6	\\\bottomrule
\end{tabular}}
\vspace{0.5pt}
\caption{SUN RGB-D: Top-$k$ retrieval accuracy for 150 queries. The retrieval candidate set contains 5,050 images. We evaluate the occurrence baseline (H), 2D relation baseline (2D), CNN baseline, the proposed hard version, proposed soft versions, and a combination between our soft version and the 2D baseline. The parameter of our model  $[x,y]$ means sampling $x$ 3D layouts and $y$ camera views for each layout. All results of our model are averaged over 5 random trials. The threshold for detection outputs is 0.5. The best is shown in \textbf{bold} and the second best is shown with \underline{underline}.}\label{tab-sun}\vskip -1em
\end{table}

The top-$k$ retrieval recalls are shown in Table~\ref{tab-sun}. In addition with the baselines, two versions of our approach are evaluated. The baselines and our hard model use bounding boxes with over $0.5$ confidence and weigh them equally, while our soft models use all bounding boxes and assign their confidences as weights in Eq.~\ref{layout-match}. The results suggest that the hard model with $5$ layout samples outperforms the occurrence baseline and is on par with the 2D baseline. Our soft models perform even better than the hard one. With increased layout samples, our approach outperforms the baselines significantly. We also evaluate a combination between our soft model and the 2D baseline by adding their normalized scores. The result suggests that such combination further boost the accuracy and that our physical model based solution is complementary to learning based approaches. 

\begin{figure*}[!t]
\centering
\subfigure[\scriptsize A picture is above a bed. A night stand is on the right side of the head of the bed. A lamp is on the night stand. Another picture is above the lamp. A dresser is on the left side of the head of the bed.]{\includegraphics[width=0.3\textwidth]{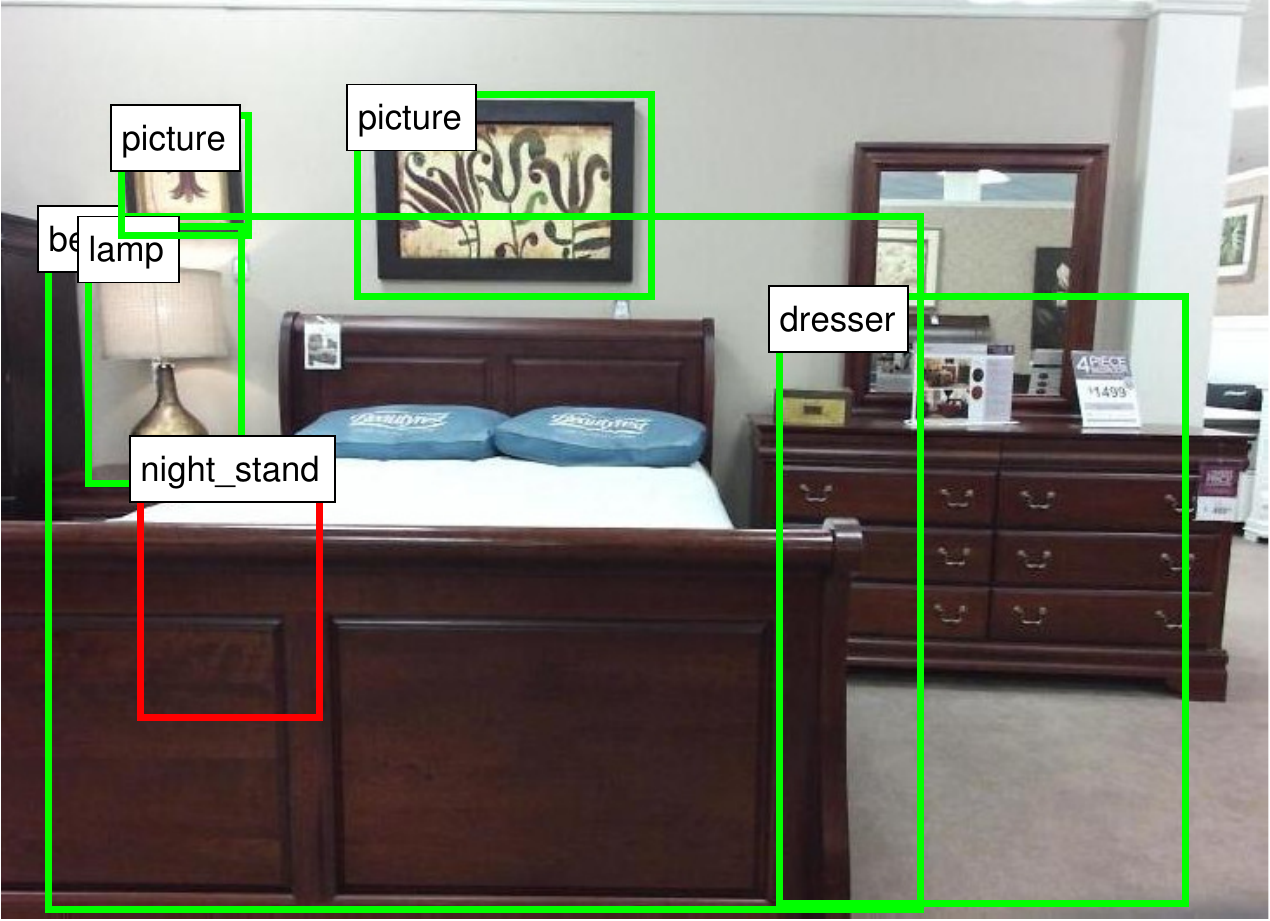}}\qquad
\subfigure[\scriptsize There is a triple sofa. The sofa is against the wall. A chair is next to the sofa. And the chair is also against the wall. Two pictures are above the sofa. And another picture is above the chair.]{\includegraphics[width=0.3\textwidth]{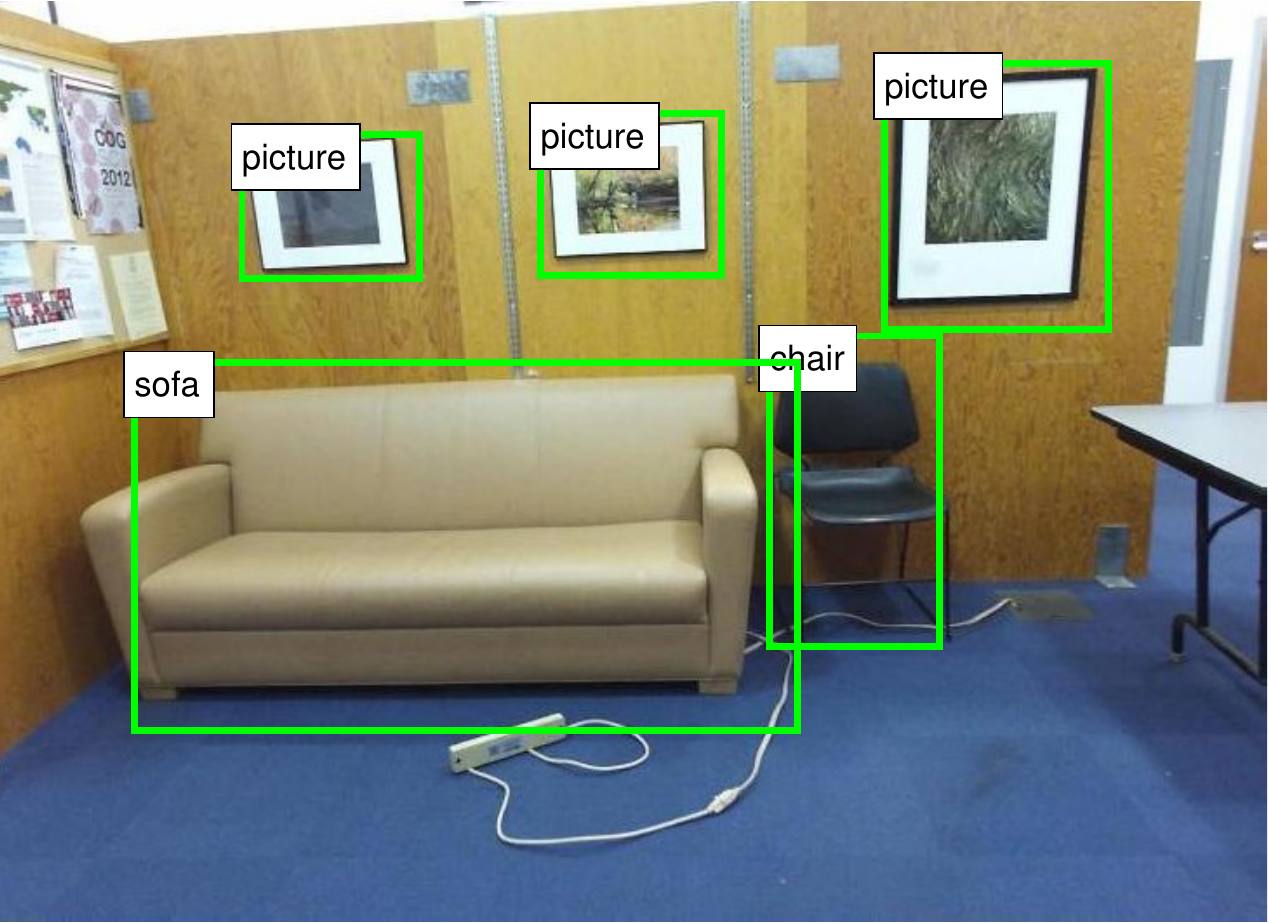}}\qquad
\subfigure[\scriptsize A chair is in front of the desk. Some boxes are on the desk. A monitor is on the desk. The desk is against the wall.]{\includegraphics[width=0.3\textwidth]{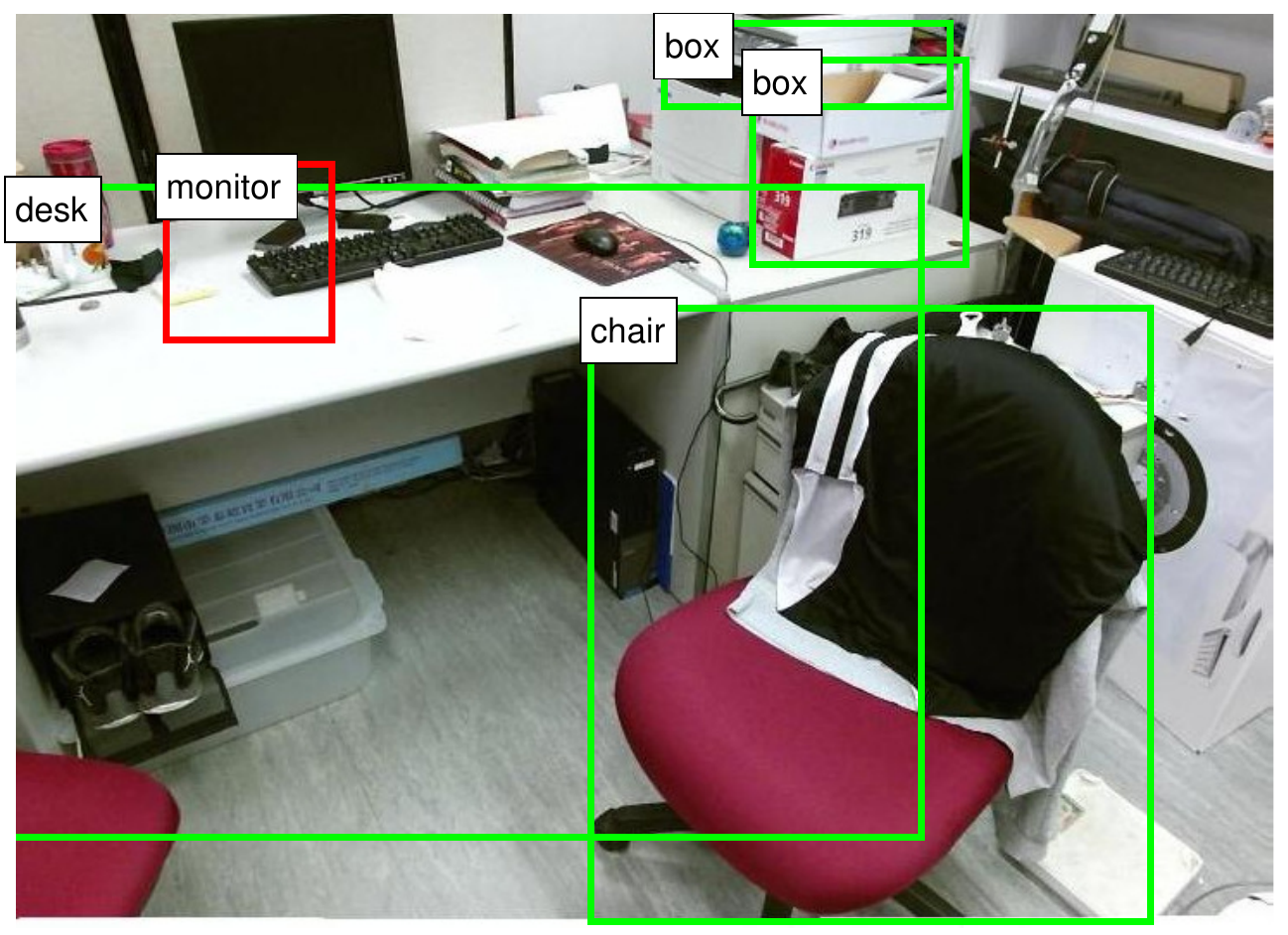}}

\caption{Matched object layouts based on our greedy 2D layout matching for three ground truth images that are ranked top 5 among all candidate images. {\color[rgb]{0,0.618,0} Green} bounding boxes are object detection outputs that match the 2D layouts generated from the text queries. {\color[rgb]{0.618,0,0} Red} bounding boxes represent a missing object (not detected by the object detector) within the expected region proposed by 2D layouts.}
 
    \label{fig:qual}
\end{figure*}

\begin{figure}[!t]
\centering
\subfigure[]{
\includegraphics[width=0.22\textwidth]{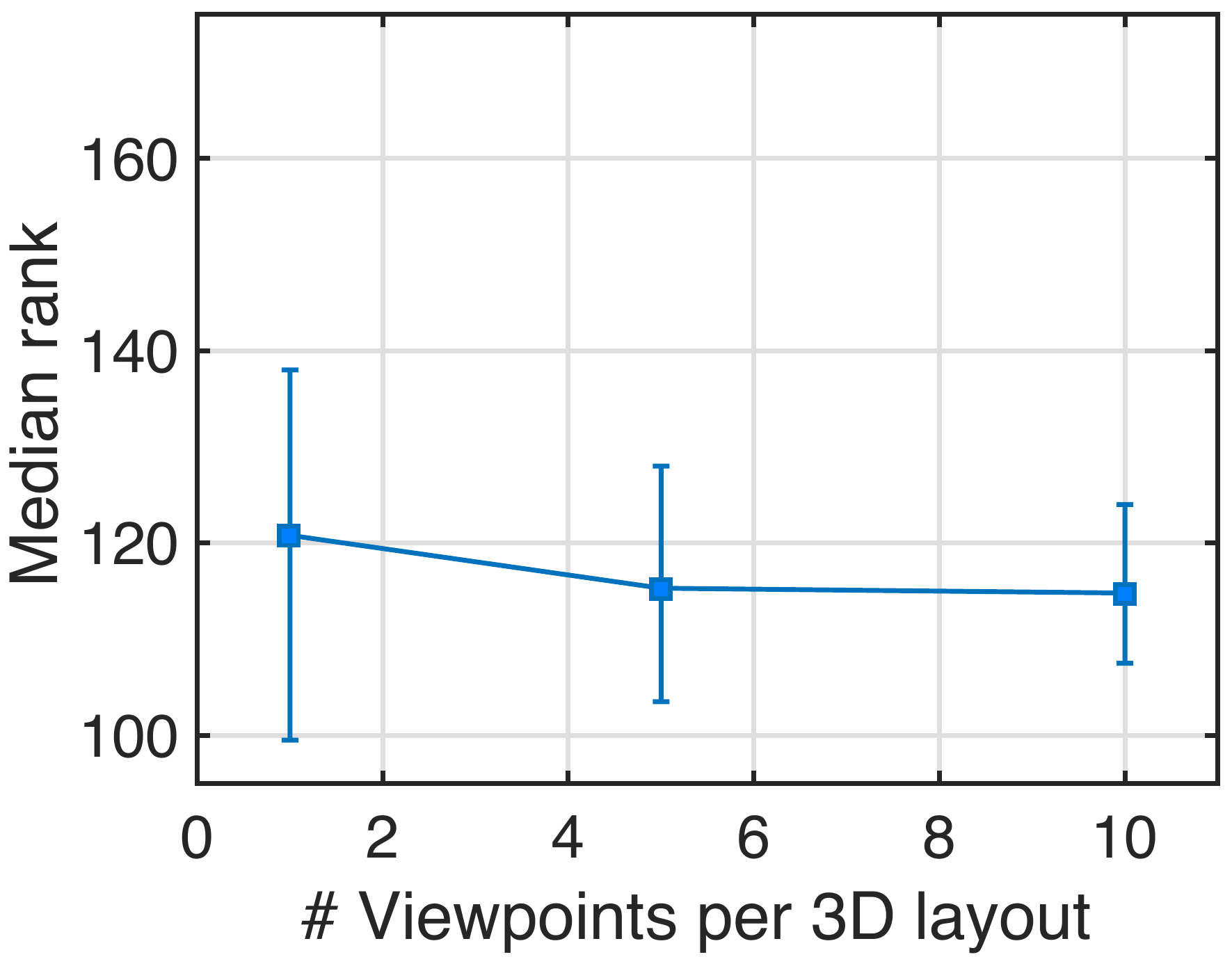}\quad
}\subfigure[]{
\includegraphics[width=0.22\textwidth]{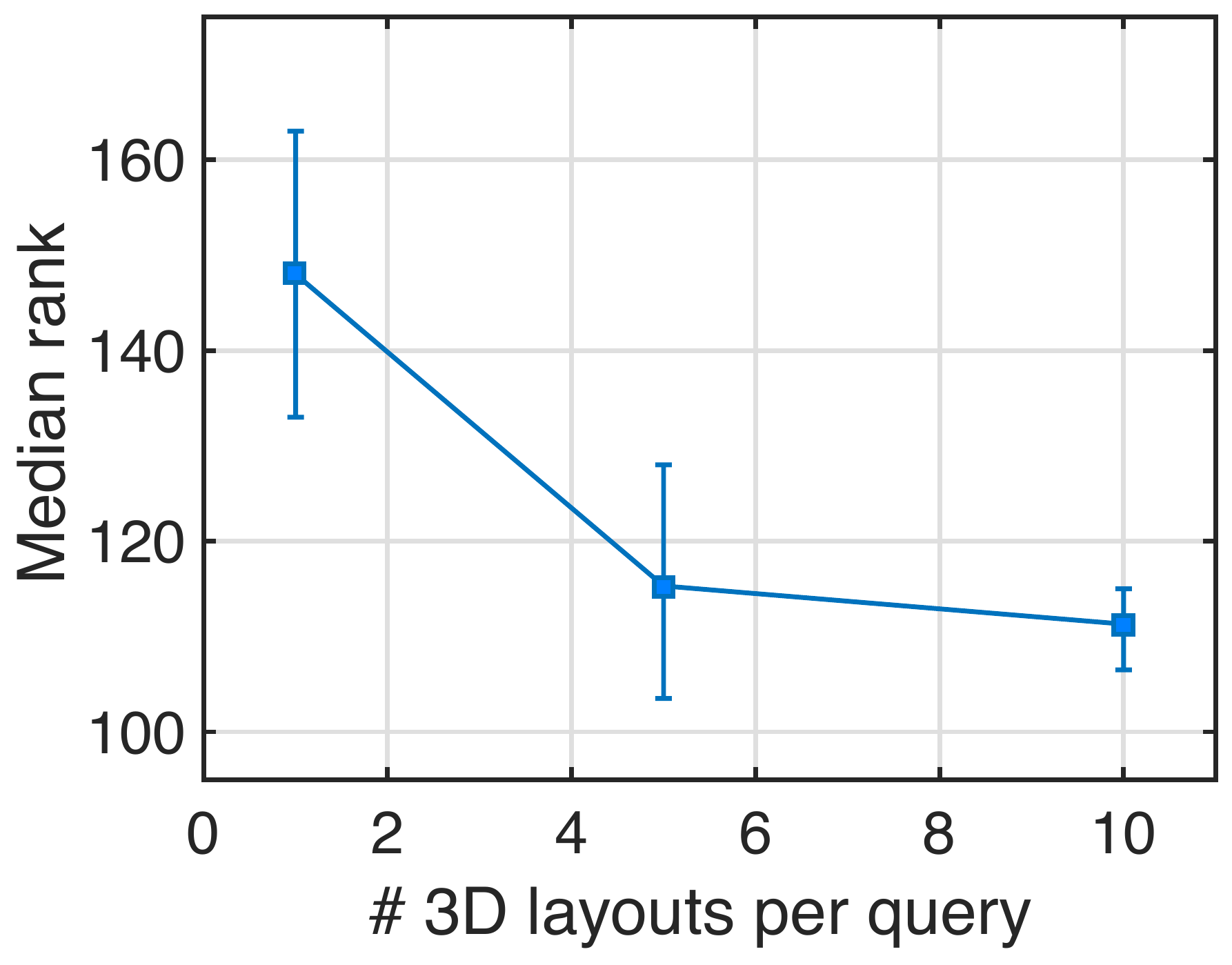}\quad
}
\caption{Influence of \# viewpoint samples and \# layout samples: (a) 5 3D layouts sampled for each query, and (b) 5 viewpoint sampled for each 3D layout. The $y$-axis is median rank of ground truths. We random 5 times for each data point. Lower is better.}
\label{fig:sample}
\end{figure}

Fig.~\ref{fig:qual} shows 3 examples whose ground truths are ranked top 5. The object bounding boxes that best match the generated 2D layouts are shown on the images. Green boxes are matched objects and red boxes are missing ones, expected in the generated 2D layout but unseen in the object detection output. The figure shows that our model has some level of tolerance on missing detections. A more interesting finding is that our model suggests potential locations for missing objects even though they could be heavily occluded.

To obtain 2D layouts, we sample 3D layouts and camera views. Fig.~\ref{fig:sample} shows how the sample size of both affects the the median rank of ground truths (keeping one and varying the other). Fig.~\ref{fig:sample} suggests that more samples generally yield better performance and the improvement saturates as the sample size increases. The improvement brought by more 3D layouts is more significant than that brought by more camera views. In addition, the performance uncertainty due to randomness decreases as the sample size increases.



\subsection{3DGP dataset with DPM detectors}
The 3DGP dataset \cite{Choi:cvpr:2013} contains $1,045$ images with three scene types: living room, bedroom and dining room. Each image is annotated with bounding boxes for 6 object categories: \textit{sofa}, \textit{table}, \textit{chair}, \textit{bed}, \textit{side table} and \textit{dining table}. Following the same protocol as in \cite{Choi:cvpr:2013}, 622 training images are used to train the furniture detectors and the remaining 423 images are used as the retrieval image database. We use pre-trained Deformable Part Models (DPM) \cite{Felzenszwalb:pami:2010} of indoor furnitures provided by the 3DGP dataset and use the thresholds in the pre-trained models to cut off false alarms. Non-maximum suppression is used to remove duplicates.

3DGP dataset is less diverse than SUN RGB-D; many images have very similar layouts. We annotated 50 unique layout descriptions which cover 222 test images. The retrieval results are shown in Table \ref{tab-3dgp-gt}. Because our method is agnostic about object detector algorithms, we split the results into two parts to separate the impact from using a specific detection algorithm: one using ground truth bounding boxes and the other using DPM detection outputs. The results suggest that our approach outperforms baseline algorithms under both bounding box settings and the improvement is independent from detector performances.




 
\begin{table}[!tb]
\centering
\resizebox{\linewidth}{!}{
\begin{tabular}{c cccc cccc}
\toprule
&\multicolumn{4}{c}{\bf w/ DPM bbox}&\multicolumn{4}{c}{\bf w/ GT bbox}\\
&H&2D&CNN&Ours&H&2D&CNN&Ours\\
\midrule
R@1&{4.0}&2.0&		4.0&\bf {4.4}&	\underline{4.0}&	\underline{4.0}	&		\underline{4.0}&{3.0}\\
R@10&10.0&14.0&		16.0&\bf {16.8}&16.0&18.0	&	14.0&\underline{20.2}\\
R@50&30.0&30.0&		30.0&\bf {31.2}&34.0&38.0	&	32.0&\underline{41.4}\\
R@100&46.0&32.0&		32.0&\bf {52.0}&64.0&66.0	& 66.0&\underline{68.0}\\\bottomrule
\end{tabular}}
\vspace{3pt}
\caption{3DGP dataset: Top-K image retrieval accuracy. Left half is based on DPM (the best is with {\bf bold}) and right half is based on ground truth bounding boxes (the best is in \underline{underline}). The results of our approach (soft[5,5]) are averaged over 10 random trials.}\label{tab-3dgp-gt}
\end{table}

  \section{Conclusion}
  We presented a general framework for retrieving images from a natural language description of the spatial layout of an indoor scene. The core component of our framework is an algorithm that generates possible 3D object layouts from text-described spatial relations and matching these layout proposals to the 2D image database. We validated our approach via the image retrieval task on two public indoor scene datasets and the result shows the possibility of generating 3D layout proposals for rigid objects and the effectiveness of our approach to matching them with images. 
 
 \vskip 1em
{\small\noindent\textbf{Acknowledgement.} This research was supported in part by funds provided from the Office of Naval Research under grant N000141612713 entitled Visual Common Sense Reasoning for Multi-agent Activity Prediction and Recognition.}

{\small
\bibliographystyle{ieee}
\bibliography{egbib}
}

\section*{\Large Supplementary materials}
\appendix
\section{Additional details about datasets}
\begin{figure}[!t]
\centering
\includegraphics[width=.45\textwidth]{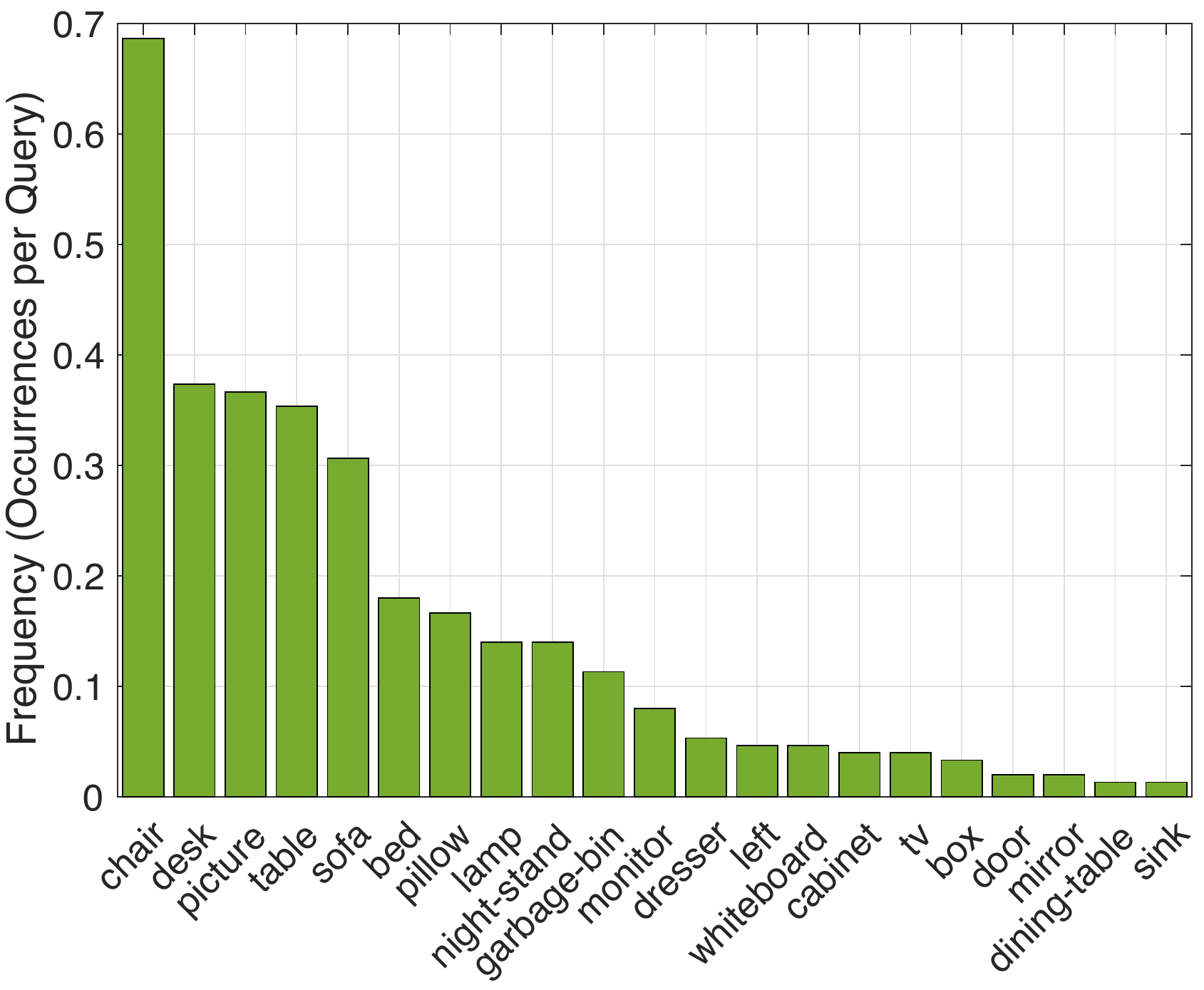}
\caption{Query statistics in SUN RGB-D evaluation: frequency (occurrences per query) of objects.}\label{fig-objstat}
\end{figure}

\begin{figure}[!t]
\centering
\includegraphics[width=.45\textwidth]{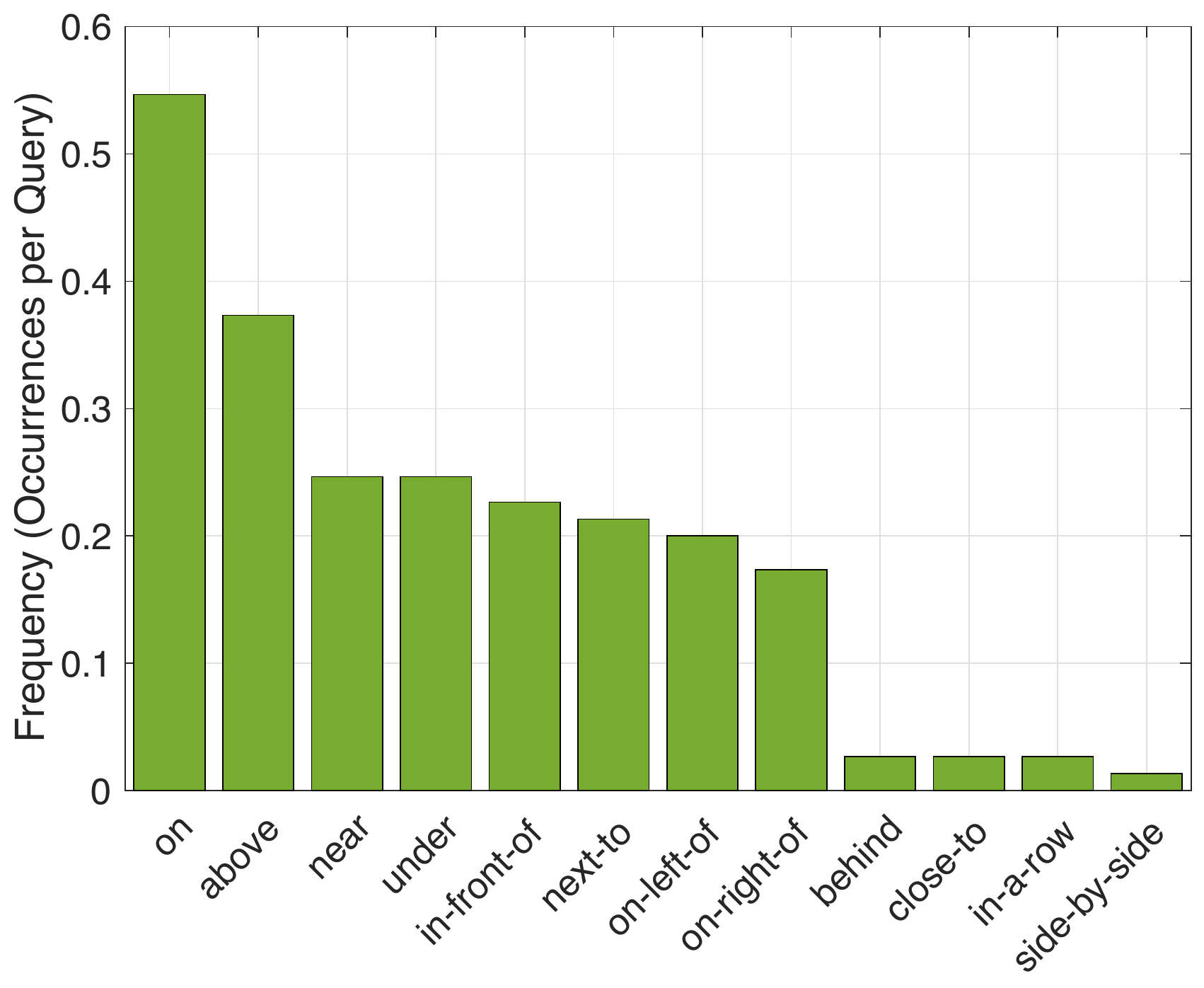}
\caption{Query statistics in SUN RGB-D evaluation: frequency (occurrences per query) of spatial relations. }\label{fig-relstat}
\end{figure}

We have 150 of the images annotated in the SUN RGB-D test dataset. We show the statistics about queries used in SUN RGB-D evaluation. In average, SUN RGB-D annotation has $4.26$ objects, $2.65$ relations, $19.85$ words and $2.69$ sentences per query. Fig. \ref{fig-objstat} shows the averaged occurrences per query of each object category and Fig. \ref{fig-relstat} shows the averaged occurrences per query of each spatial relation category. The object and relation categories are sorted in the descending order w.r.t. the frequency.

\begin{figure}[!t]
\centering
\subfigure[]{\includegraphics[width=.24\textwidth]{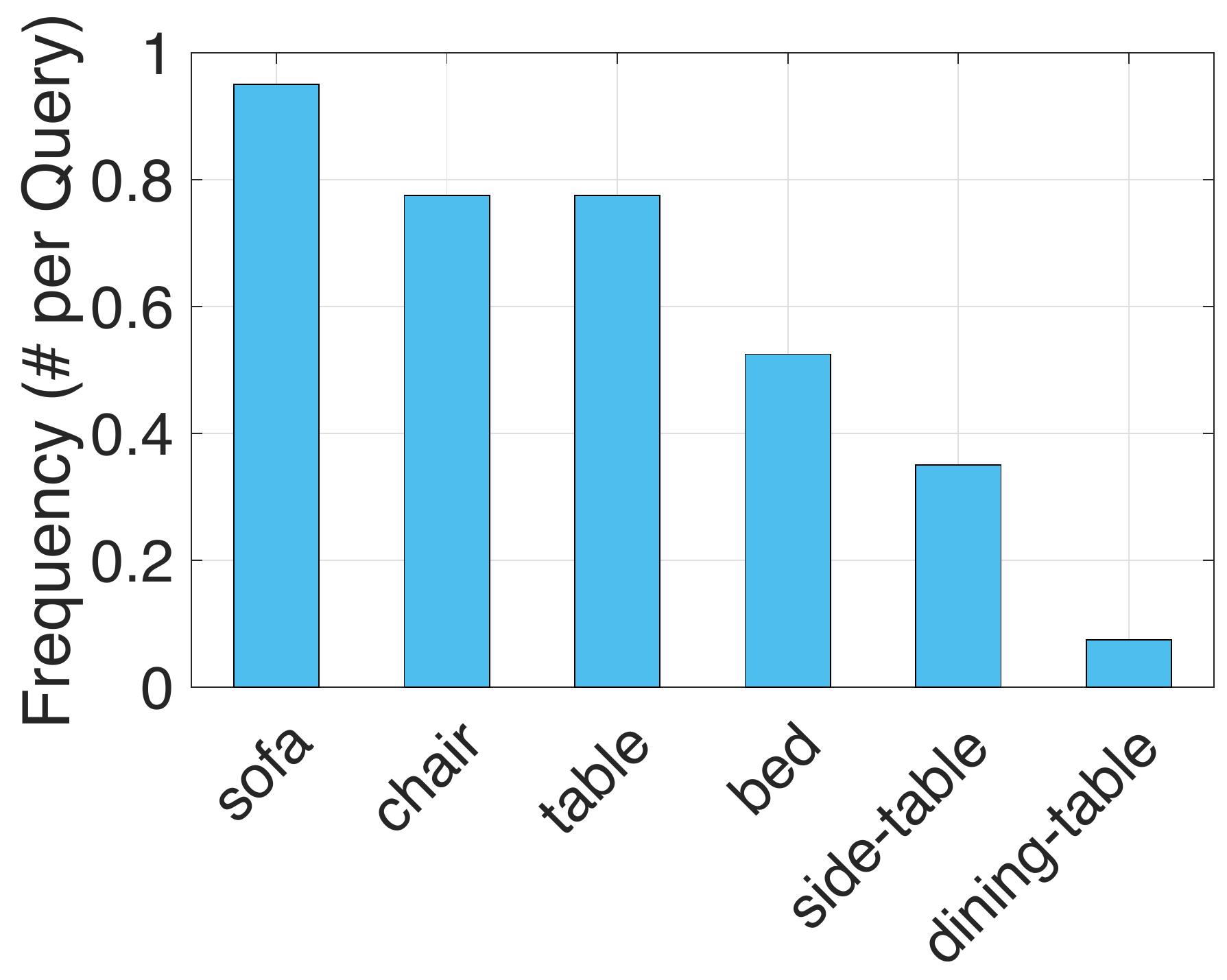}\label{fig-objstat-3dgp}}\subfigure[]{
\includegraphics[width=.24\textwidth]{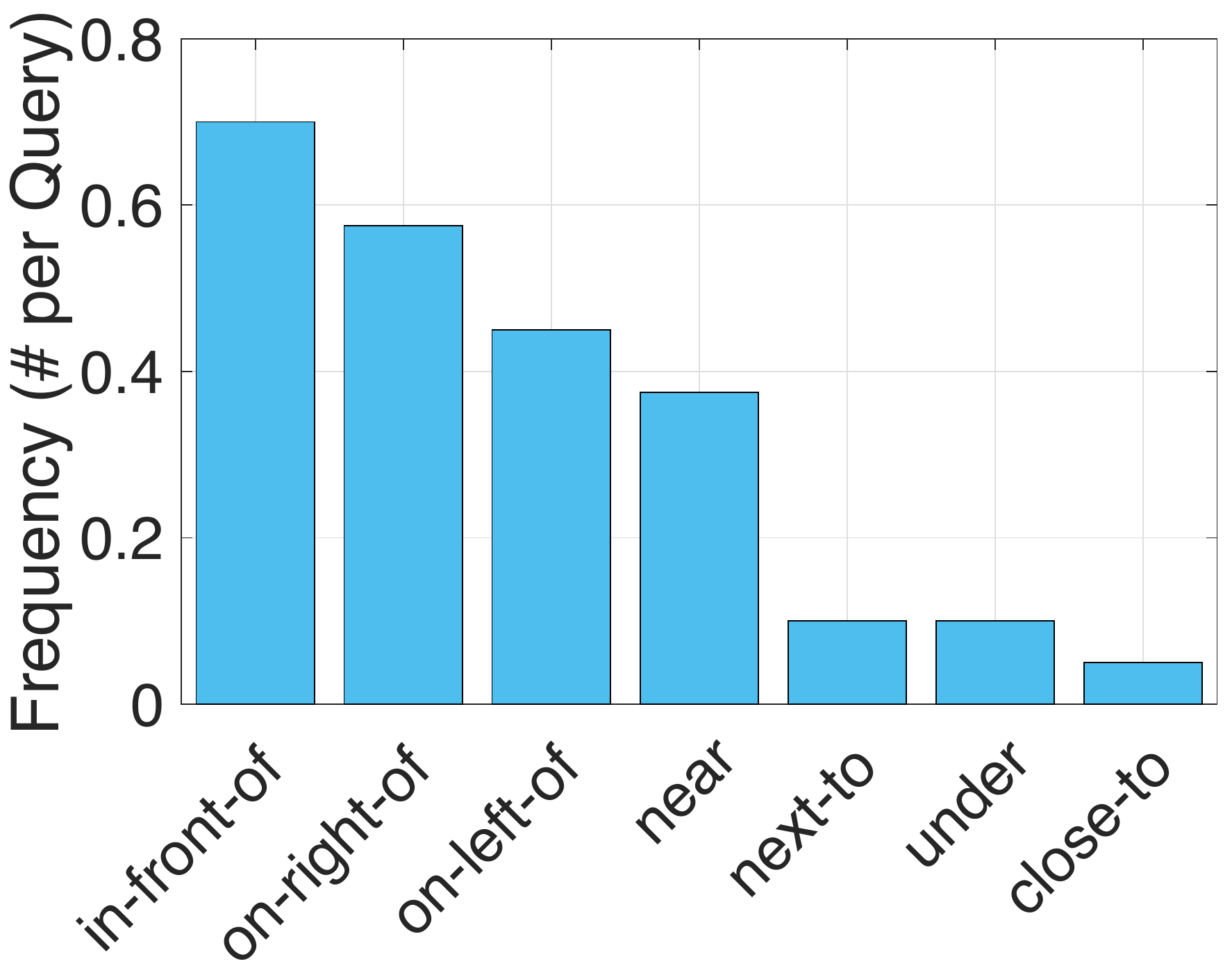}\label{fig-relstat-3dgp}}
\caption{Query statistics in 3DGP evaluation:  (a) Frequency (occurrences per query) of objects, and (b) frequency (occurrences per query) of spatial relations.}\label{fig-stat-3dgp}
\end{figure}

In average, 3DGP annotation has $3.06$ objects, $1.94$ relations,  $17.06$ words and $1.94$ sentences per query. Similarly, we show the object category and spatial relation frequencies in Fig. \ref{fig-stat-3dgp}. Different from the statistics of SUN RGB-D where spatial relation \texttt{on} has the highest frequency, spatial relations in 3DGP are mostly horizontal. This is because, for 3DGP, we only have DPM detectors for 6 furniture categories and all of them are on the floor.

\section{Additional qualitative retrieval results}
\begin{figure*}[!t]
\centering
\includegraphics[width=0.97\textwidth]{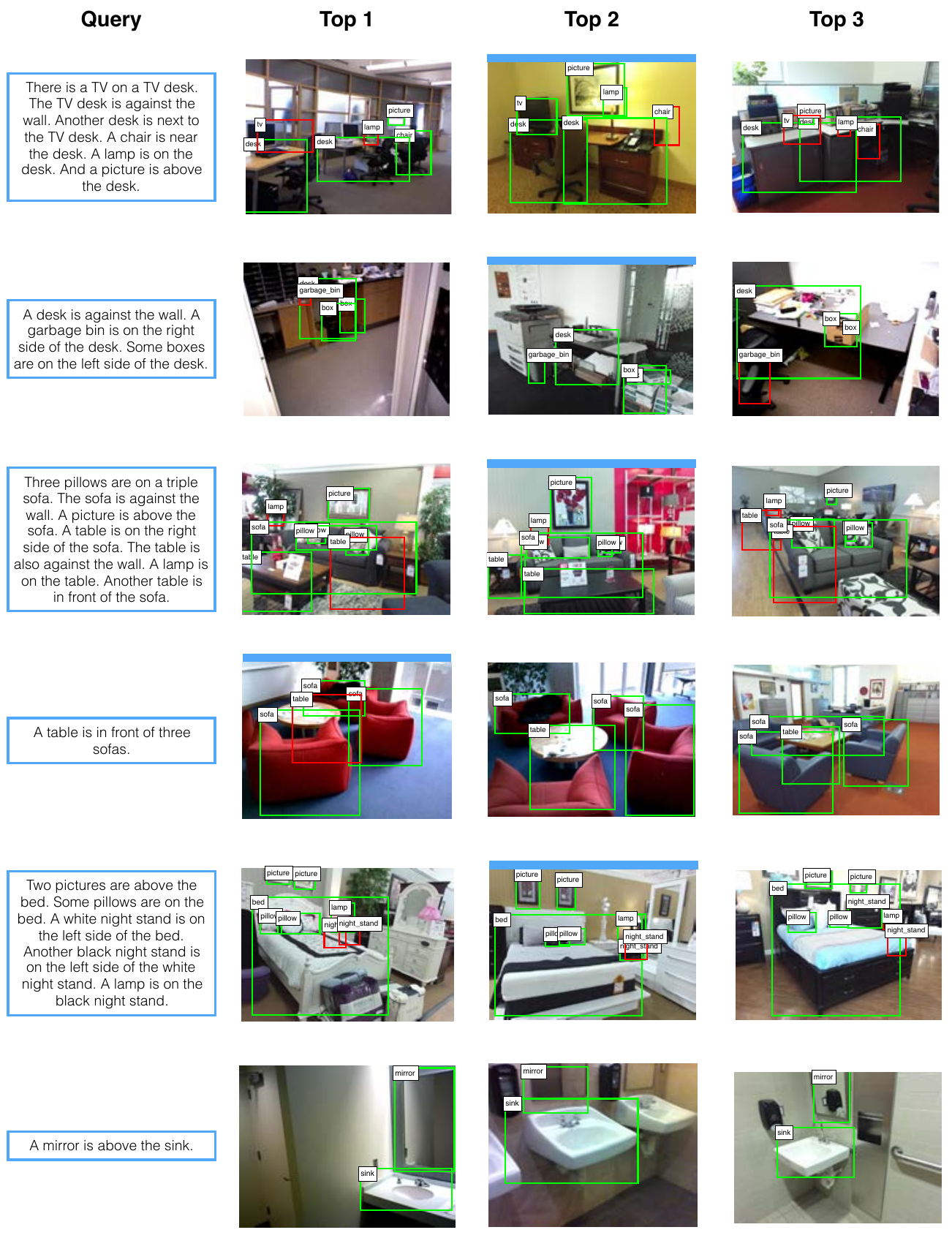}
\caption{Top 3 retrieved images in SUN RGB-D. Ground truth images appear with {\color[rgb]{0,0.2,0.8}\bf blue} bars on top. {\color[rgb]{0,0.618,0} Green} bounding boxes are detection outputs matching the generated 2D layouts. {\color[rgb]{0.618,0,0} Red} boxes are missing objects (not detected) w.r.t. the expectation of generated 2D layouts.}\label{add-retrieval}
\end{figure*}

\begin{figure*}[!t]
\centering
\includegraphics[width=0.85\textwidth]{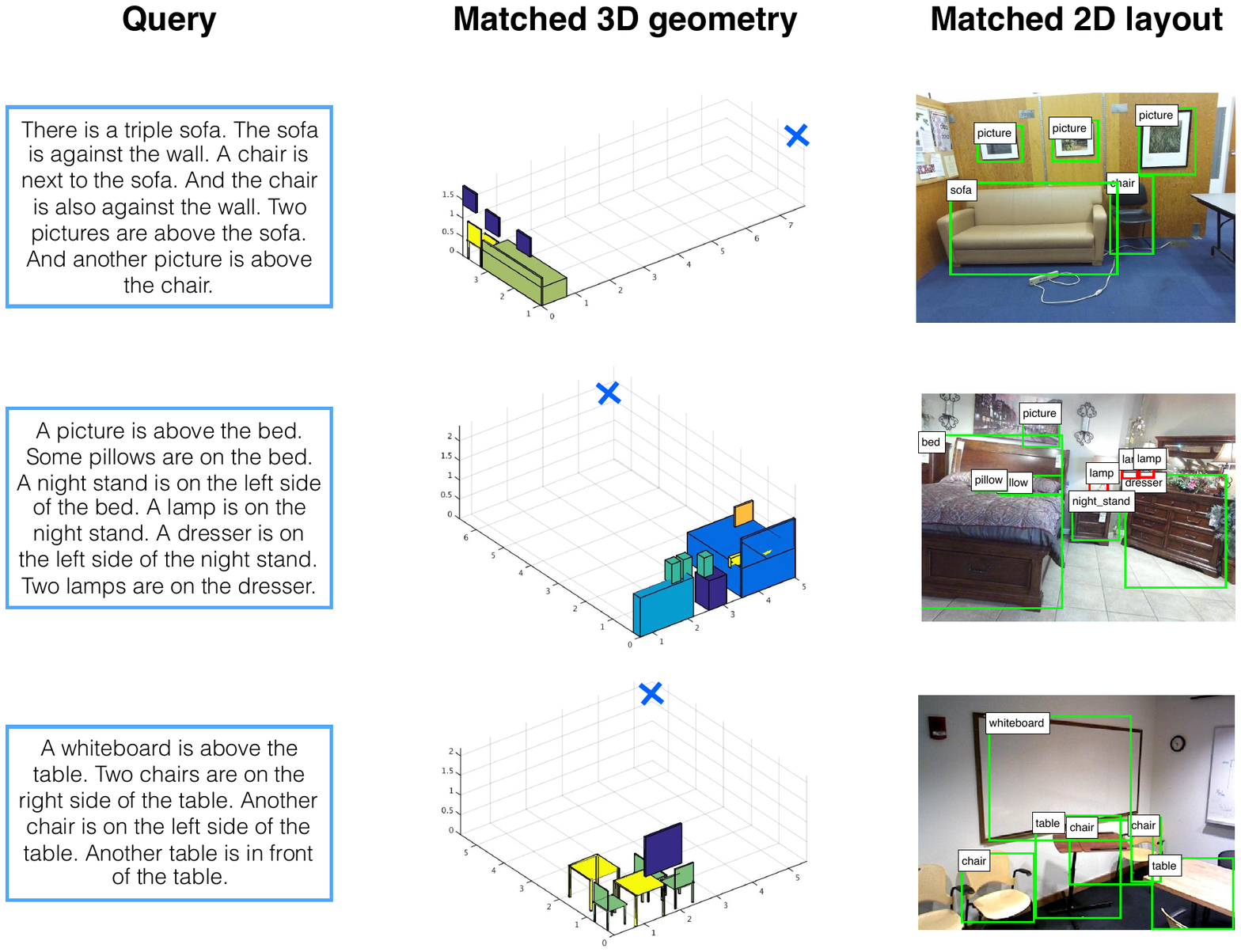}

\caption{Matched 3D and 2D layouts based on our 2D layout matching for three ground truth images in SUN RGB-D. {\color{blue}Blue} crosses represent camera locations. {\color[rgb]{0,0.618,0} Green} bounding boxes are object detection outputs that match the 2D layouts generated from the text queries. {\color[rgb]{0.618,0,0} Red} bounding boxes represent a missing object (not detected by the object detector) within the expected region proposed by 2D layouts.}\label{add-match}
\end{figure*}

We provide additional results in SUN RGB-D for top-3 retrievals in Fig.~\ref{add-retrieval}. From left to right, the columns in the figure show the query (text description of the spatial relationships), the retrieved image ranked 1st, the retrieved image ranked 2nd and the retrieved image ranked 3rd. The ground truth image is shown with a blue bar on its top. Although it happens rare in this evaluation, there are cases when there are images other than the ground truth that meet the descriptions of the query (e.g., the last example in Fig.~\ref{add-retrieval}). Qualitative results with matched 3D layout are shown in Fig.~\ref{add-match} for three example images. The figure shows the 3D layouts with camera location corresponding to the best matched 2D spatial layouts (from five layout samples). 

%
%
%

\section{Learned 2D spatial relationships in baseline}
\begin{figure*}[!htb]
\centering
\subfigure[on]{
\includegraphics[width=.23\textwidth]{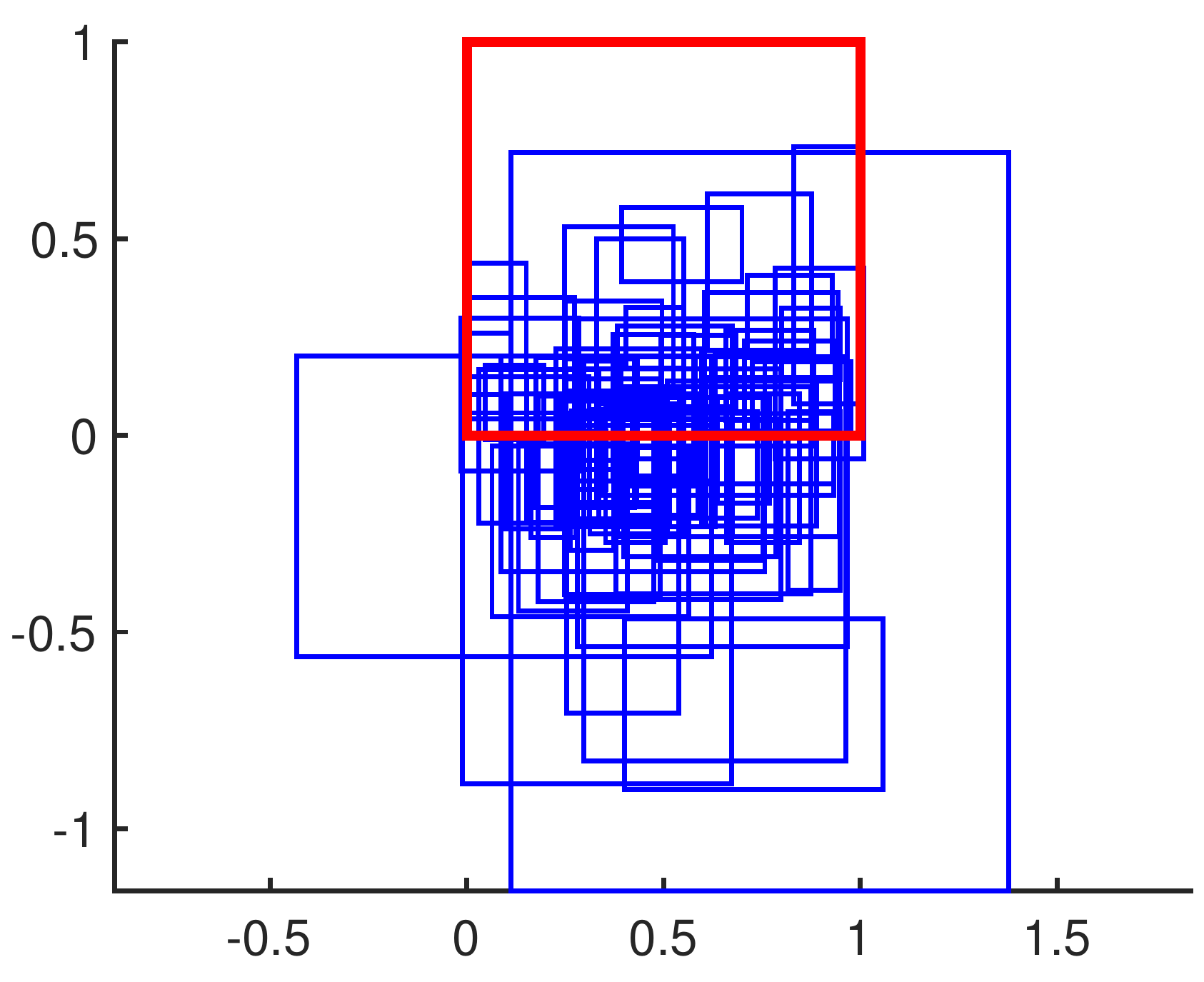}
}
\subfigure[under]{
\includegraphics[width=.23\textwidth]{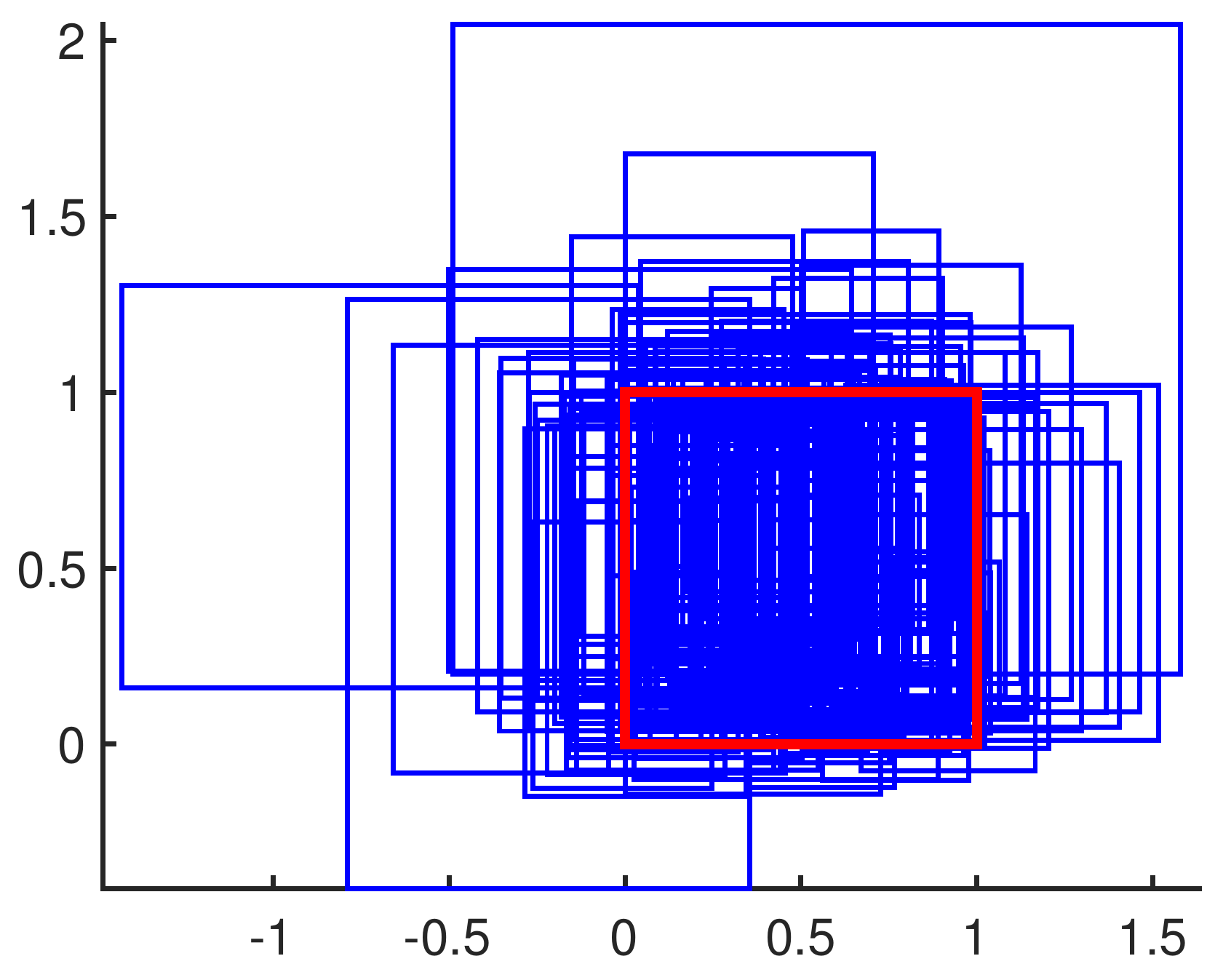}
}
\subfigure[above]{
\includegraphics[width=.23\textwidth]{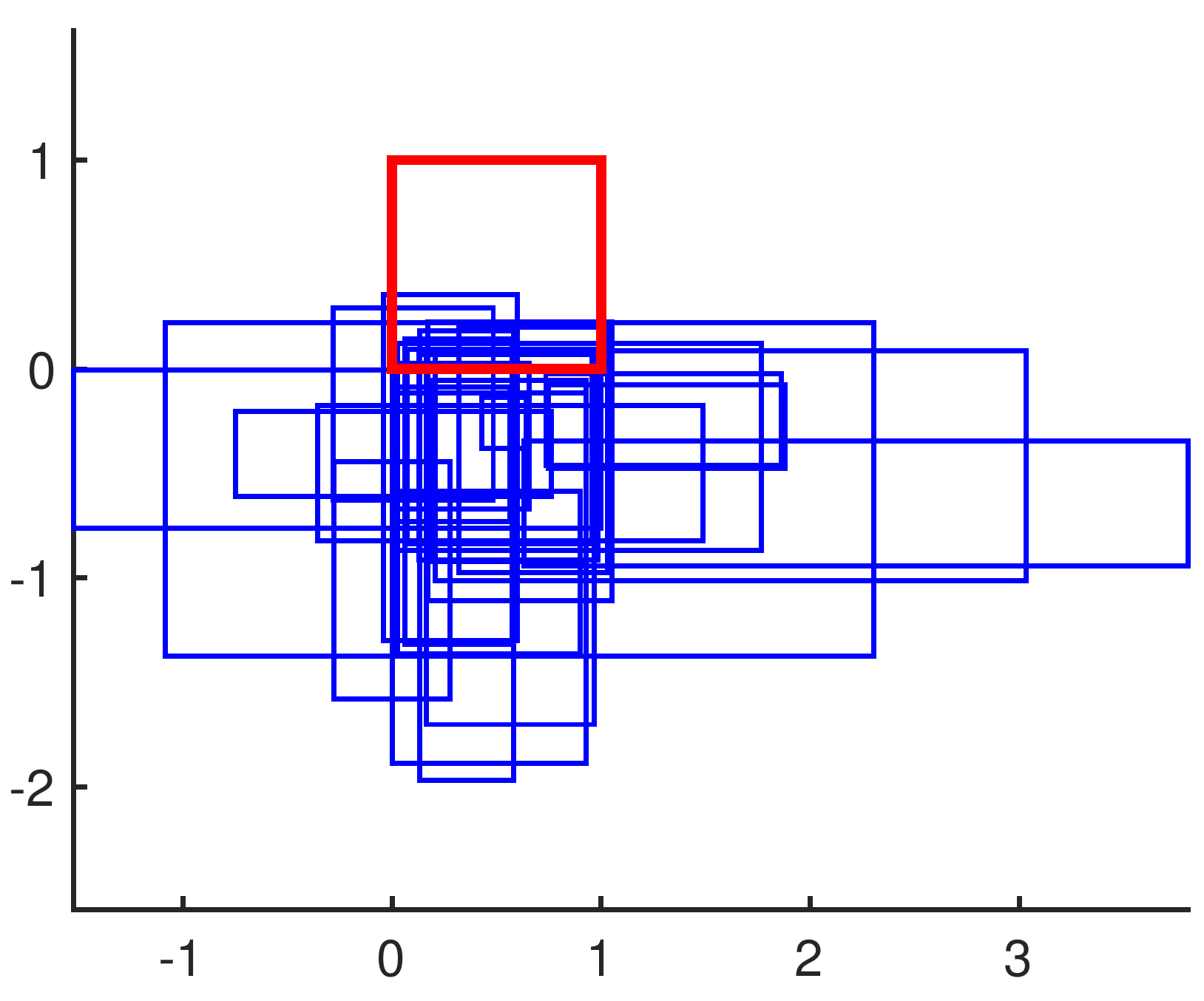}
}
\subfigure[near]{
\includegraphics[width=.23\textwidth]{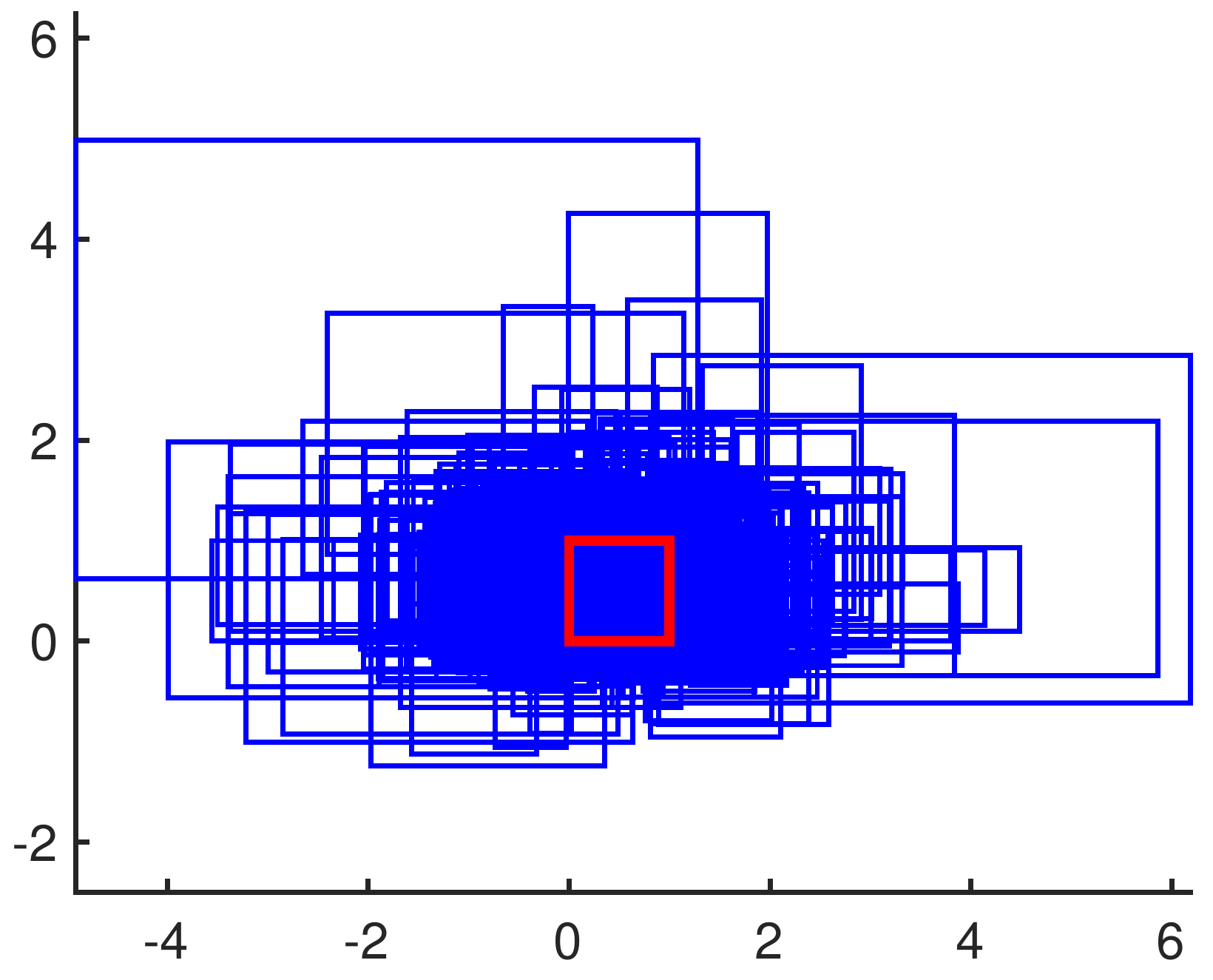}
}
\subfigure[left]{
\includegraphics[width=.23\textwidth]{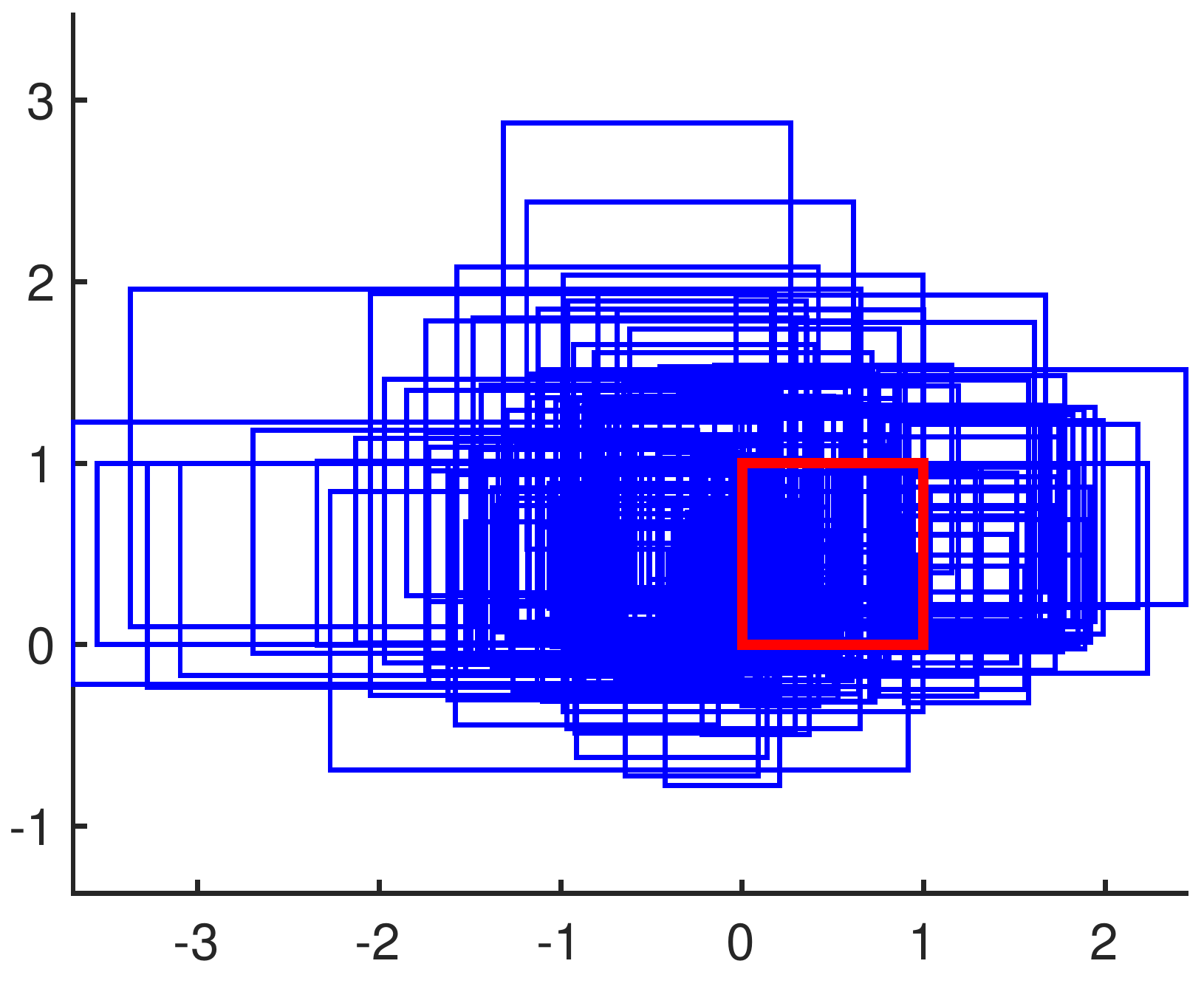}
}
\subfigure[right]{
\includegraphics[width=.23\textwidth]{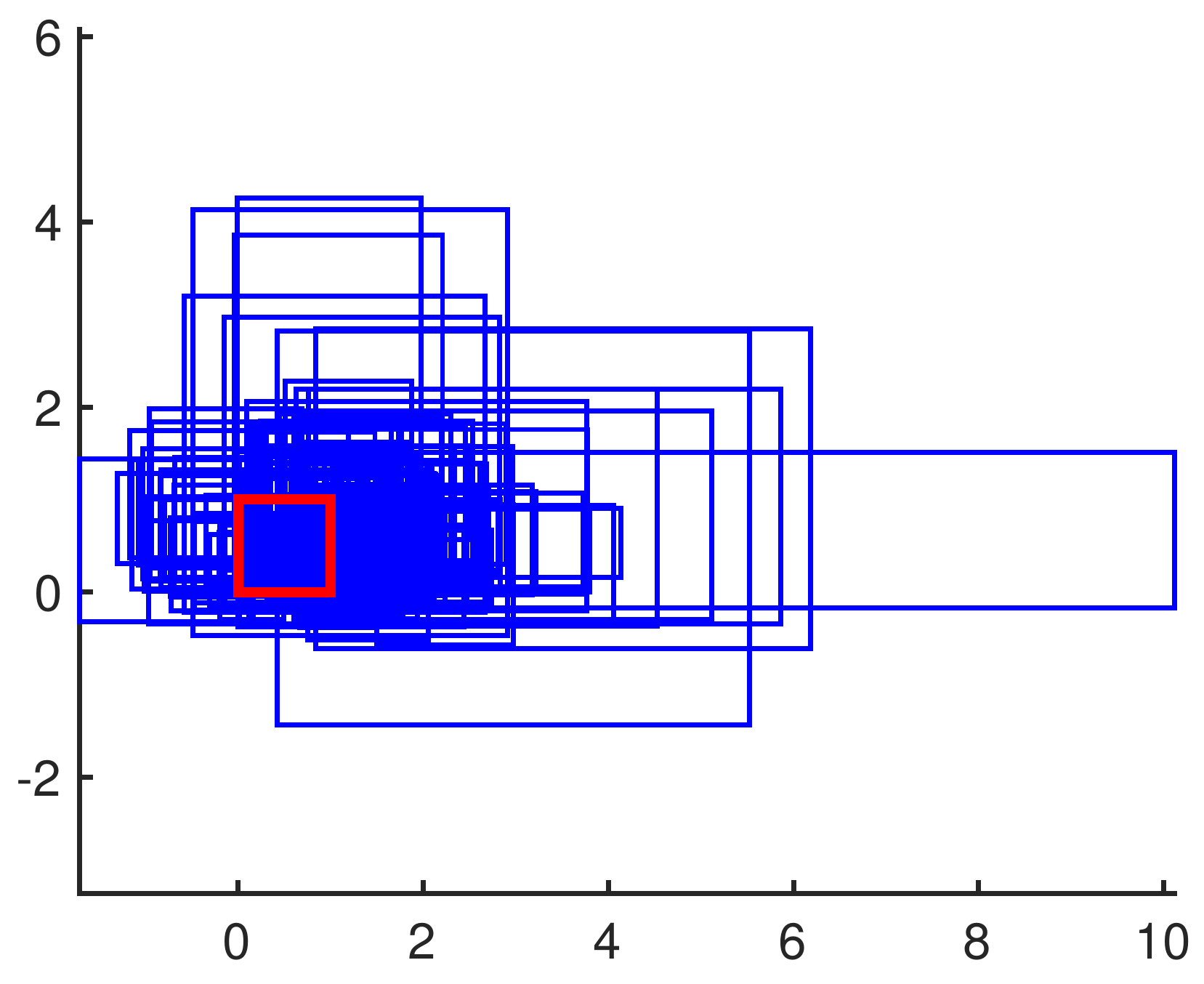}
}
\subfigure[front]{
\includegraphics[width=.23\textwidth]{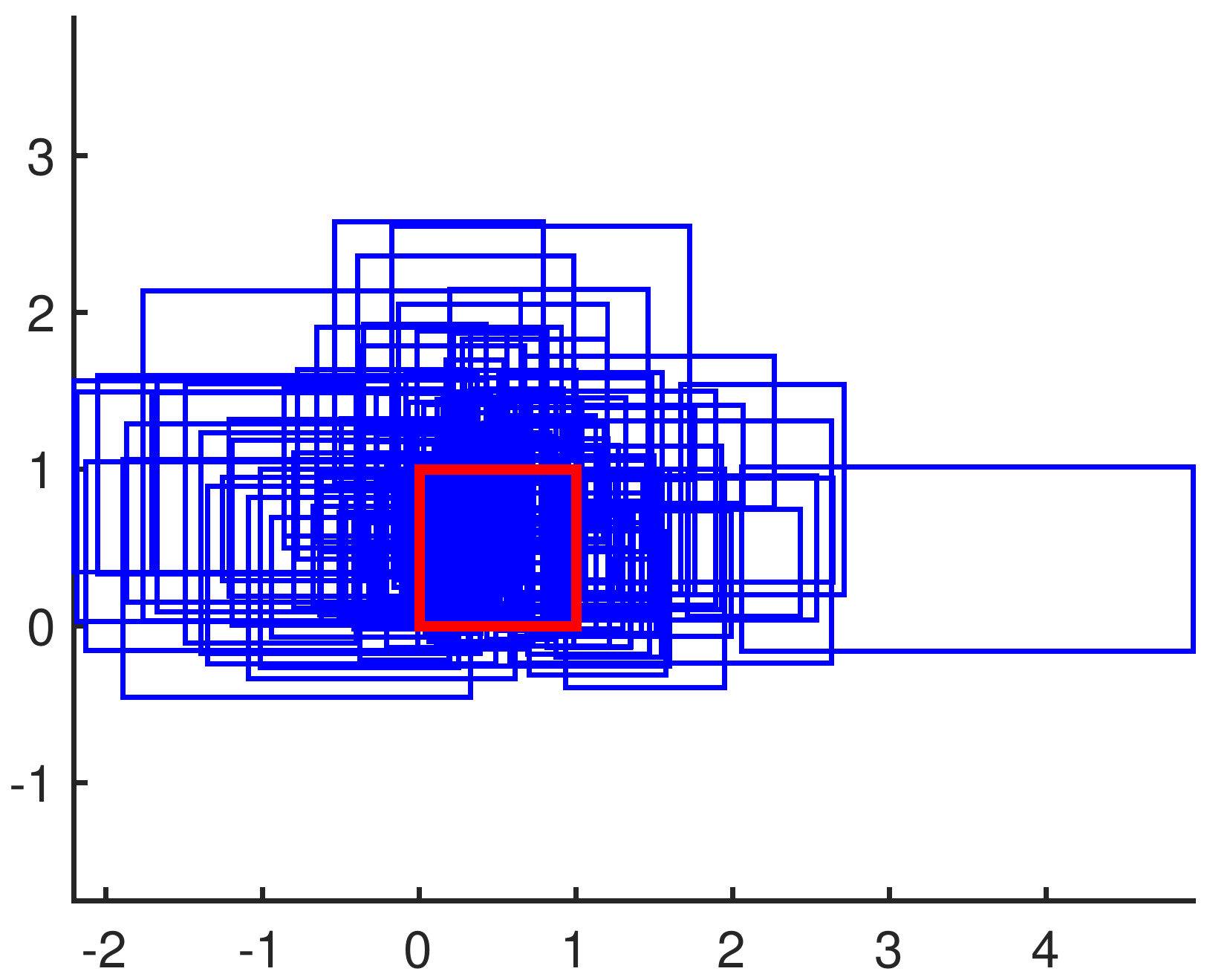}
}
\subfigure[behind]{
\includegraphics[width=.23\textwidth]{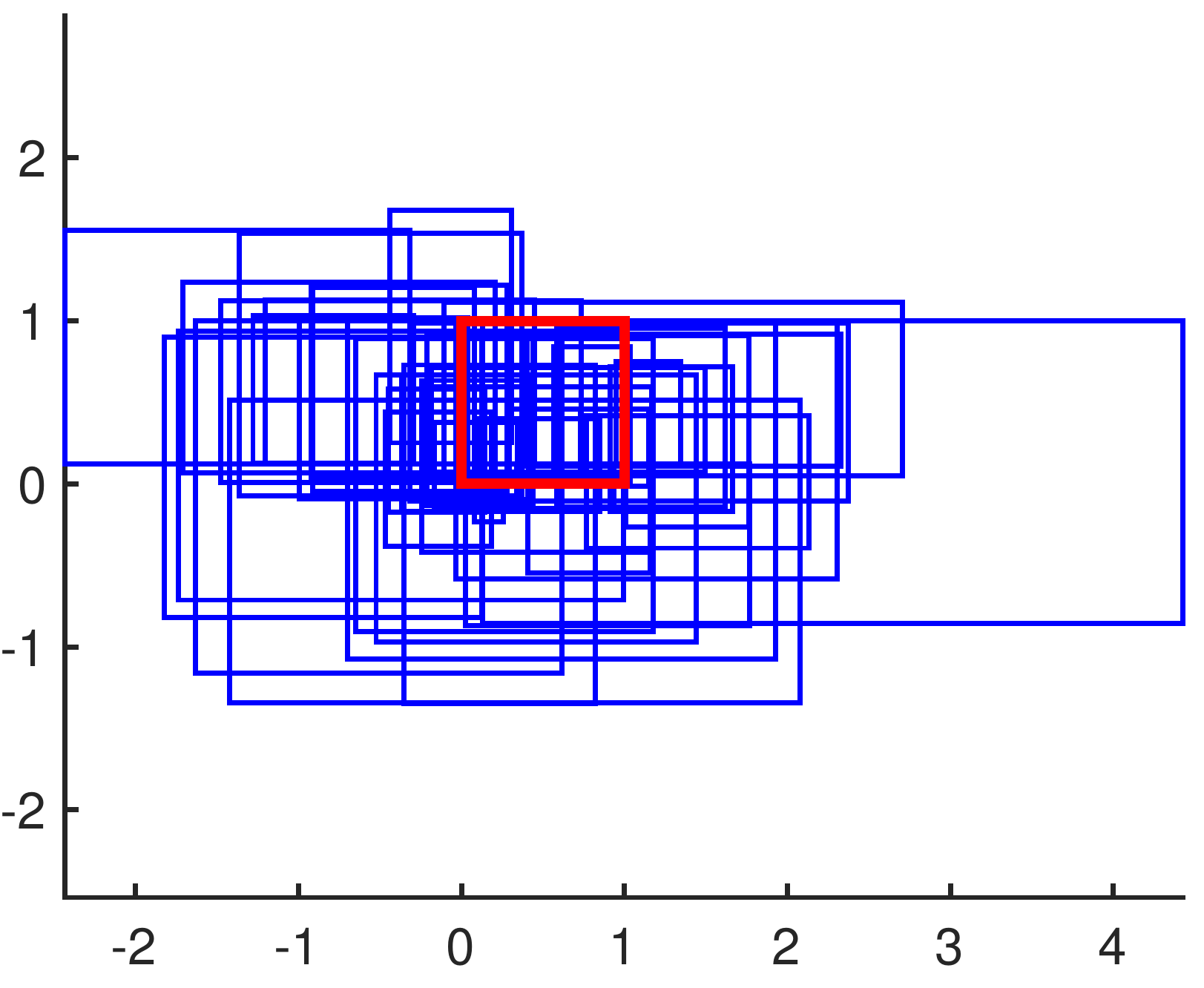}
}
\caption{Learned distribution of 2D spatial relations in \textbf{subject-relation-object} relationships. {\color{red} Red} bounding boxes represent the subject  and blue bounding boxes represent the sampled objects in the annotations corresponding to each relation. The subject is normalized to $1\times 1$ squares (with bottom-left $(0,0)$ and top-right $(1,1)$) and all objects are rescaled with the same normalization factors in $x$-$y$ coordinates.}\label{2d}
\end{figure*}

The learned distributions of 2D spatial relationships in the nearest neighbor baseline algorithm are shown in Fig. \ref{2d}. The figure shows the relationship between the subject and the object (in \textit{subject-relation-object} relationships) w.r.t. all eight atomic spatial relations (other relations are built upon these atomic relations). For each relationship, the annotated bounding boxes of each pair of subjects and objects are normalized (rescaled in both $x-$ and $y-$ coordinates) so that the subject bounds to a $1\times 1$ square with bottom left $(0,0)$ and top right $(1, 1)$. All of the normalized relation annotations are visualized in the figure. The nearest neighbor classifier we used in the baseline algorithm is achieved by computing the intersection over union (\textsc{IOU}) scores of the normalized object bounding boxes.

\end{document}